\PassOptionsToPackage{dvipsnames}{xcolor}

\documentclass{article}
\usepackage{bbm}
\usepackage[utf8]{inputenc} 
\usepackage[T1]{fontenc}    
\usepackage{hyperref}       
\usepackage{url}            
\usepackage{booktabs}       
\usepackage{amsfonts}       
\usepackage{nicefrac}       
\usepackage{microtype}      
\usepackage{lipsum}		
\usepackage{graphicx}
\usepackage{float}
\usepackage{natbib}
\usepackage{doi}
\usepackage{xcolor}
\usepackage[normalem]{ulem}

\usepackage{color}
\usepackage{lineno}
\usepackage{amsmath}
\usepackage{xcolor}

\usepackage{soul}
\usepackage{arxiv}
\usepackage{url}
\usepackage{bm}
\newcommand\myshade{85}
\colorlet{mylinkcolor}{RoyalBlue}
\colorlet{mycitecolor}{violet}
\colorlet{myurlcolor}{YellowOrange}

\hypersetup{
  linkcolor  = mylinkcolor!\myshade!black,
  citecolor  = mycitecolor!\myshade!black,
  urlcolor   = myurlcolor!\myshade!black,
  colorlinks = true,
}

\usepackage{times}
\usepackage{latexsym}

\usepackage[T1]{fontenc}

\usepackage[utf8]{inputenc}

\usepackage{microtype}

\usepackage{booktabs}
\usepackage{graphicx}
\usepackage{array}
\usepackage{multirow}
\usepackage{longtable}
\usepackage{CJKutf8}
\usepackage{float}
\usepackage{microtype}
\usepackage{multirow}
\usepackage{booktabs}
\usepackage{graphicx}
  
\usepackage{amssymb}
\usepackage{url}
\usepackage{hyperref}
\usepackage{amsmath}
\usepackage{caption}
\usepackage[ruled,vlined,linesnumbered]{algorithm2e}
\usepackage{subfigure}
\usepackage{makecell}
\usepackage{ulem}
\usepackage{algorithmic}
\usepackage[misc]{ifsym}
\usepackage{lipsum}
\usepackage{colortbl}

\usepackage{tikz}
\usetikzlibrary{shapes.geometric}

\usepackage{palatino} 

\title{Read to Play (\includegraphics[width=0.031\linewidth]{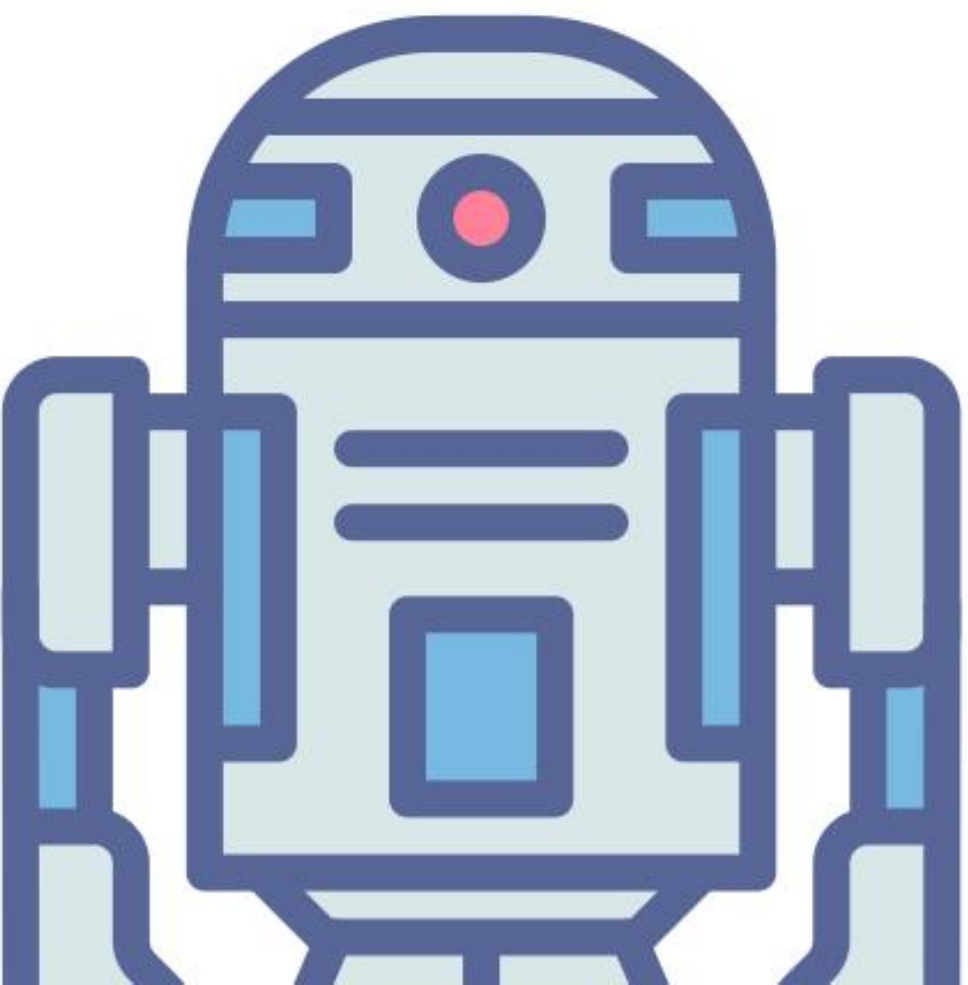} R2-Play): Decision Transformer with Multimodal Game Instruction}

\newcommand\blfootnote[1]{%
\begingroup
\renewcommand\thefootnote{}\footnote{#1}%
\addtocounter{footnote}{-1}%
\endgroup
}

\author{ 
    \textbf{\textsuperscript{2}Yonggang Jin $^{\ast}$},\space
    \textbf{\textsuperscript{1,4,5}Ge Zhang $^{\ast}$$^{\dag}$},\space
    \textbf{\textsuperscript{2}Hao Zhao $^{\ast}$},\space
    \textbf{\textsuperscript{2}Tianyu Zheng},\space
    \textbf{\textsuperscript{2}Jarvi Guo},\space
    \\
    \textbf{\textsuperscript{6}Liuyu Xiang},\space
    \textbf{\textsuperscript{6}Shawn Yue},\space
    \textbf{\textsuperscript{6}Stephen W. Huang},\space
    \textbf{\textsuperscript{2}Zhaofeng He $^{\dag}$},\space
    \textbf{\textsuperscript{3}Jie Fu $^{\dag}$},\space
    \\
{\small 
\textsuperscript{1} Multimodal Art Projection Research Community;
}
{\small 
\textsuperscript{2} Beijing University of Posts and Telecommunications;
} \\
{\small 
\textsuperscript{3} HKUST;
} 
{\small 
\textsuperscript{4} University of Waterloo;
}
{\small 
\textsuperscript{5} Vector Institute;
}
{\small 
\textsuperscript{6} Harmony.AI;
}
}

\begin{document}
\maketitle

\vspace{-7ex}
\begin{center}
     \url{https://r2-play.github.io}
\end{center}
\vspace{3ex}

\begin{abstract}
Developing a generalist agent is a longstanding objective in artificial intelligence. 
Previous efforts utilizing extensive offline datasets from various tasks demonstrate remarkable performance in multitasking scenarios within Reinforcement Learning.
However, these works encounter challenges in extending their capabilities to new tasks.
Recent approaches integrate textual guidance or visual trajectory into decision networks to provide task-specific contextual cues, representing a promising direction. 
However, it is observed that relying solely on textual guidance or visual trajectory is insufficient for accurately conveying the contextual information of tasks.
This paper explores enhanced forms of task guidance for agents, enabling them to comprehend gameplay instructions, thereby facilitating a ``read-to-play'' capability. 
Drawing inspiration from the success of multimodal instruction tuning in visual tasks, we treat the visual-based RL task as a long-horizon vision task and construct a set of multimodal game instructions to incorporate instruction tuning into a decision transformer. 
Experimental results demonstrate that incorporating multimodal game instructions significantly enhances the decision transformer's multitasking and generalization capabilities. 
Our code and data are available at \url{https://github.com/ygjin11/R2-Play}.
\end{abstract}

\blfootnote{$^{\ast}$ These authors contributed equally.}
\blfootnote{$^{\dag}$ Corresponding Authors.}

\section{Introduction}

Creating a generalist agent that can accomplish diverse tasks is an enduring goal in artificial intelligence.
Recently, ~\citet{mgdt22, generalist} showcase exceptional performance in multitasking scenarios within Reinforcement Learning (RL) using extensive offline datasets that cover a wide range of decision tasks. 
However, despite the significant achievements, these models still face challenges in adapting to novel tasks, primarily due to insufficient task-specific knowledge and contextual information.
Thus, developing a generalist agent capable of accomplishing diverse tasks while demonstrating adaptability to novel tasks remains a formidable obstacle.

\begin{figure}[!htbp]
\centering
\includegraphics[width=0.6\linewidth]{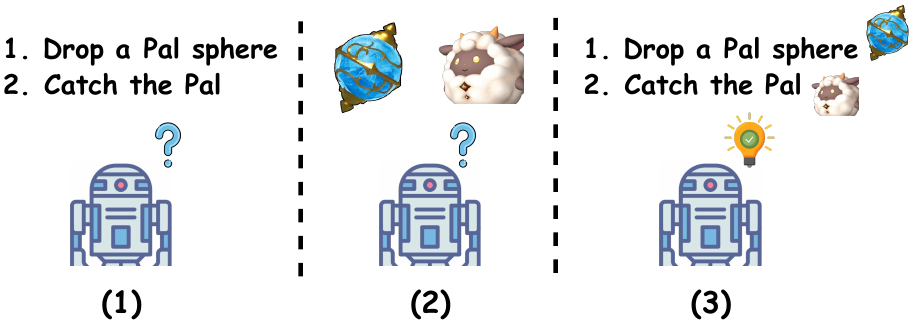}
\caption{Imagine an agent learning to play Palworld (a Pok\'emon-like game). 
(1) The agent exhibits confusion when only relying on textual guidance.
(2) The agent is confused when presented with images of a Pal sphere and a Pal.  
(3) The agent understands how to catch a pet through multimodal guidance, which combines textual guidance with images of the Pal sphere and Pal.
} 
\label{fig_teaser}
\end{figure}

The recent advancement in integrating textual guidance ~\citep{reward22, controller23, planner22} or visual trajectory ~\citep{hyperdt23, icdt23, groot23} into a single decision-making agent presents a potential solution.
This line of research provides task-specific context to guide the agent.
Although textual guidance and visual trajectory each offer advantages, they also have distinct limitations: (1) Textual guidance lacks visually-grounded information, which diminishes its expressiveness for decision-making tasks based on visual observations~\citep{palm23, groot23}; (2) Without clear task instructions, deriving an effective strategy from a visual trajectory is extremely difficult, which is similar to people's difficulty understanding player intentions when watching game videos without explanations.
The complementary relationship between textual guidance and visual trajectory suggests their combination enhances guidance effectiveness, as illustrated in Figure \ref{fig_teaser}.
As a result, this paper aims to \textit{develop an agent capable of adapting to new tasks through multimodal guidance}.

\vspace{-0.25cm}

Similar endeavors are undertaken in the field of multimodal models. 
~\citet{otter23, vi23} integrate instruction tuning into multimodal models, enhancing their visual task performance. 
Drawing inspiration from the success of multimodal instruction tuning in visual tasks, we treat the visual-based RL task as a long-horizon vision task, aiming to integrate it into the RL field.
Given the extensive variety of games and the availability of expert human gameplay videos in the Atari game set (as detailed in Section \ref{setup}), it is chosen as the experimental environment for our research. 
We construct a set of \textbf{M}ultimodal \textbf{G}ame \textbf{I}nstruction (\textbf{MGI}) to provide multimodal guidance for agents.
The MGI set comprises thousands of game instructions sourced from approximately 50 diverse Atari games, designed to provide a detailed and thorough context. 
Each instruction entails a 20-step trajectory, labeled with corresponding textual language guidance.  
The construction of this multimodal game instruction set aims to empower agents to read game instructions for playing various games and adapting to the new ones.

To augment agents' multitasking and generalization capabilities via multimodal game instructions, leveraging a large, diverse offline dataset for pretraining is vital. This approach is supported by the success of instruction tuning in large language models~\citep{instructgpt, gpt3} and large multimodal models~\citep{otter23, vi23}.
The Decision Transformer (DT), an innovative integration of transformers with reinforcement learning (RL), demonstrates superior multitasking capabilities in RL, attributed to its large-scale pretraining ~\citep{dt21, mgdt22}.
This advancement establishes DT as a leading approach in the domain of large-scale pretraining for RL. 
Expanding upon the DT framework, we introduce the \textbf{D}ecision \textbf{T}ransformer with \textbf{G}ame \textbf{I}nstruction (\textbf{DTGI}). DTGI is pretrained using Multimodal Game Instructions and a comprehensive offline dataset, designed to simulate the agent's ability to follow game instructions. 
Simultaneously, we introduce a novel design called \emph{SHyperGenerator}, which facilitates knowledge sharing between training and unseen game tasks.
Experimental results show that integrating multimodal game instruction into DT leads to a substantial improvement in multitasking and generalization performance. 
Furthermore, multimodal instruction outperforms textual language and visual trajectory, demonstrating its superior capacity to provide detailed and comprehensive context. 
Our contributions are summarized as follows:
\begin{itemize}
    \item We construct a set of  \textbf{M}ultimodal \textbf{G}ame \textbf{I}nstruction (\textbf{MGI}) to include thousands of game instructions for decision control. Each instruction corresponds to a trajectory, labeled with its corresponding language guidance to offer detailed and contextual understanding.
    \item We propose the \textbf{D}ecision \textbf{T}ransformer with \textbf{G}ame \textbf{I}nstruction (\textbf{DTGI}) that enhances DT-based agents' ability to comprehend game instructions during gameplay. Additionally, we propose a novel design named \emph{SHyperGenerator} to enable knowledge sharing between training and unseen game tasks.
    \item Experimental results demonstrate that the incorporation of multimodal game instructions significantly improves the multitasking and generalization of the DT, surpassing the performance achieved through textual language and visual trajectory in isolation.
\end{itemize}

\section{Preliminary}
In this section, we present the backgrounds of the Decision Transformer and Hypernetwork.

\subsection{Decision Transformer}
The Decision Transformer, proposed by ~\citet{dt21}, combines sequence modeling and Transformers with Reinforcement Learning, deviating from traditional RL approaches that primarily focus on creating policies or value functions.
This model utilizes a GPT-style language model architecture ~\citep{gpt2}, to analyze historical sequences of states, actions, and accumulated rewards to predict future actions. 
A key feature of the Decision Transformer is the integration of the \textbf{return-to-go} (rtg) into its input, orienting the model's attention towards actions expected to maximize rewards.
The rtg at a timestep \(t\) within a trajectory denoted as \((s_1, a_1, \hat{R_1},\ldots, s_t, a_t, \hat{R_t}, \ldots, s_T, a_T, \hat {R_T})\), where \(s_t\), \(a_t\), and \(\hat {R_t}\) represent states, actions, and rtgs respectively, is calculated by the accumulation of the discounted future rewards: $\hat{R_T} = \sum_{k=t}^{T} \gamma^{k-t} r_k$. Here, \(\gamma\) is the discount factor, typically ranging from 0 to 1, representing the importance of future rewards.

\subsection{Hypernetwork}
Hypernetworks~\citep{hypernetL17} employ an auxiliary network to dynamically generate the parameters of a target network.
In this framework, a target network \(\mathbf{f_{\theta}}(x)\) produces output for an input \(x\), and its weights \(\mathbf{\theta}\) are generated by an independent hypernetwork \(\mathbf{h_{\phi}}(z)\) in response to a contextual input \(z\), expressed as $\mathbf{\theta} = \mathbf{h_{\phi}}(z)$. 
Consequently, the target network's function is redefined as  $\mathbf{f}(x|z) = \mathbf{f}_{\mathbf{h_{\phi}}(z)}(x)$. 
This architecture enables the target network to adapt its parameters dynamically for a variety of tasks or input scenarios without the need for retraining from the ground up. Such flexibility is beneficial in areas like multitask, few-shot, and zero-shot learning.
The hypernetwork \(\mathbf{h_{\phi}}(z)\), with its own parameters \(\mathbf{\phi}\), can be simultaneously optimized with the target network by minimizing a mutual loss function, enhancing collaborative learning of both target network parameters and hypernetwork parameters.

\section{Method}
\subsection{Problem Formulation}
In the multitask offline dataset, each game task, denoted as \(T_k\), within the training set \(\mathbf{S_{\text{train}}}\) is associated with its corresponding dataset \(D_k\). 
The dataset \(D_k\) is constructed from trajectories generated using an unspecified policy \(\mathbf{\pi}\). 
To enrich contextual information, a game instruction set \(I_k\) is linked to each task \(T_k\) in this paper. 
Consequently, the training set is defined as \(\mathbf{S_{\text{train}}} = \{(T_k, D_k, I_k) \mid k = 1, \ldots, E\}\), where \(E\) signifies the total count of training game tasks. 
During the training phase, this training set \(\mathbf{S_{\text{train}}}\) is utilized to train the DT. 
Similarly, the testing set is defined as \(\mathbf{S_{\text{test}}} = \{(T_u, I_u) \mid u = 1, \ldots, W\}\), with \(W\) representing the number of unseen game tasks. Notably, there is no dataset included in \(\mathbf{S_{\text{test}}}\). 
During the evaluation phase, the \textbf{In-Distribution} (ID) performance of the model is evaluated using game tasks from \(\mathbf{S_{\text{train}}}\), while the \textbf{Out-of-Distribution} (OOD) performance is assessed using game tasks from \(\mathbf{S_{\text{test}}}\).

\subsection{Game Instruction Construction}
\begin{figure}[!htbp]
\centering
\includegraphics[width=0.65\linewidth]{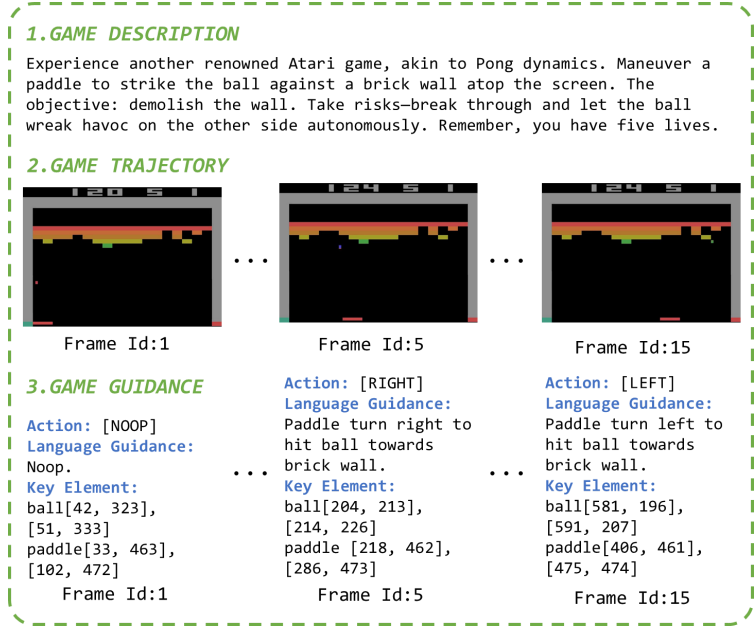}
\caption{An illustrative example of game instructions. Each instruction consists of three sections: game description, game trajectory, and game guidance (including action, language guidance, and the position of key elements)} 
\label{fig_instruction}
\end{figure}

In the multimodal community, considerable attention is devoted to the integration of instruction tuning within multimodal models  ~\citep{1mit23, llama-adapter23, otter23, mmrlhf23, mmic23}.
These efforts aim to enhance the performance of multimodal models in performing specific visual tasks by incorporating language instructions.
Drawing upon the insights derived from these efforts, it is imperative to explore the potential of leveraging multimodal game instructions to augment the capabilities of RL agents, especially in the context of visual-based RL tasks as long-horizon visual tasks. 
We construct a set of Multimodal Game Instruction (MGI) to apply the benefits of instruction tuning to DT.
An example of multimodal game instruction is presented in Figure \ref{fig_instruction}.

We introduce a systematic approach for formulating game instructions to explore the incorporation of instruction tuning in RL agents.
Initially, we provide ChatGPT ~\citep{chatgpt} with a thorough overview of the game, encompassing its action space.
Following that, we instruct ChatGPT to generate a detailed description for each action, focusing on highlighting key elements.
Secondly, we collect gameplay videos featuring expert human players. Subsequently, we downsample and partition these videos into N segments, each segment containing 20 frames.
Lastly, we annotate the actions and positions of key elements, represented as [a, b], [c, d]. [a, b] denotes the lower-left point of the element bounding box, while [c, d] denotes the upper-right point.

\subsection{Decision Transformer with Game Instruction}
\begin{figure}[!ht]
\centering
\includegraphics[width=0.50\linewidth]{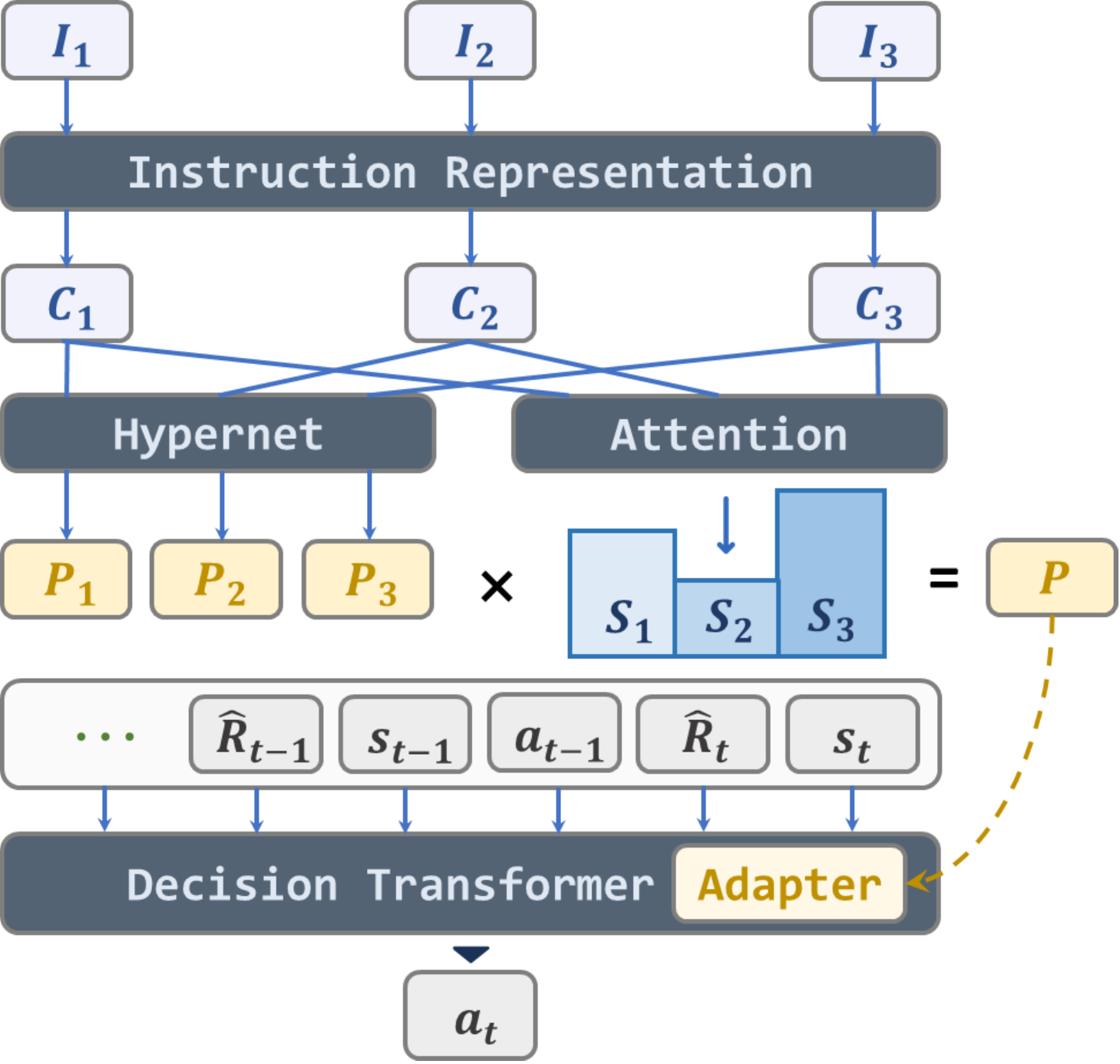}
\caption{Model architecture of Decision Transformer with Game Instruction (DTGI).
Firstly, we undertake the representation of multimodal instructions (Section \ref{instruction}) 
Secondly, we calculate importance scores for each instruction in the Instruction set (Section \ref{importance}). 
Finally, We propose a novel design named \emph{SHyperGenerator} to integrate game instructions into DT. 
N instruction generates n module parameters through hypernetworks.
The module parameters are weighted based on the importance score of the instruction, and then utilized as adapter parameters (Section \ref{dt}).
}
\label{fig_method}
\end{figure}

The current section introduces the Decision Transformer with Game Instruction (DTGI), a DT model that integrates multimodal game instructions, as depicted in Figure \ref{fig_method}. 
The formulation of DTGI is expressed as follows: 
\begin{equation}
action = \mathrm{DTGI}(trajectory, instruction)
\end{equation}
In Section \ref{instruction}, we elaborate on the representation of multimodal instructions.
Section \ref{importance} assesses the significance of individual instructions within the instruction set.
Subsequently, Section \ref{dt} delves into the integration of multimodal instructions into the Decision Transformer.

\subsubsection{Multimodal Instruction Representation}
\label{instruction}
To provide contextual details for game tasks, we construct a comprehensive set of multimodal game instructions. Specifically, for a given game $G$, we present a instruction set denoted as $I_G$, organized as follows:
\begin{equation}
I_G = \{I_1, \ldots, I_\tau, \ldots, I_n\}
\end{equation}
\begin{equation}
I_\tau = \{d_\tau, f^1_\tau, g^1_\tau, \ldots, f^i_\tau, g^i_\tau, \ldots, f^m_\tau, g^m_\tau\}
\end{equation}
$I_G$ comprises $n$ game instructions labeled as $\{I_1, \ldots, I_\tau, \ldots, I_n\}$. $I_\tau$ consists of a game description ($d_\tau$) and $m$ pairs of $(f^i_\tau, g^i_\tau)$. Here, $f^i_\tau$ refers to the i-th frame in the game trajectory$\{f^1_\tau, ..., f^i_\tau, ..., f^m_\tau\}$, and $g^i_\tau$ represents the corresponding game guidance for $f^i_\tau$.

To align the visual trajectory with language guidance, we employ the feature extractor from the frozen CLIP model ~\citep{clip}. This model includes two encoders: $\mathrm{CLIP_{img}}$ for image encoding and $\mathrm{CLIP_{txt}}$ for text encoding. The $\mathrm{CLIP_{img}}$ model extracts image features of $\{f^1_\tau, ..., f^i_\tau, ..., f^m_\tau\}$, while the $\mathrm{CLIP_{txt}}$ model extracts text features of the game description $d_\tau$ and the game guidance $\{g^1_\tau, ..., g^i_\tau, ..., g^m_\tau\}$ as follows:
\begin{equation}
\begin{split}
f^1_\tau, \cdots, f^m_\tau = \mathrm{CLIP_{img}}(f^1_\tau, \cdots, f^m_\tau)
\end{split}
\end{equation}
\begin{equation}
\begin{split}
d_\tau, g^1_\tau, \cdots, g^m_\tau = \mathrm{CLIP_{txt}}(d_\tau, g^1_\tau, \cdots, g^m_\tau)
\end{split}
\end{equation}

To capture temporal information from the instructions, we use two encoders: $\mathrm{Encoder_f}$ and $\mathrm{Encoder_g}$, composed of attention blocks facilitating the extraction of temporal information from the sequences $\{f^1_\tau, ..., f^i_\tau, ..., f^m_\tau\}$ and $\{g^1_\tau, ..., g^i_\tau, ..., g^m_\tau\}$, respectively. This extraction results in $f_\tau$ and $g_\tau$.
\begin{equation}
f_\tau = \mathrm{Encoder_f}(\{f^1_\tau, ..., f^i_\tau, ..., f^m_\tau\})
\end{equation}
\begin{equation}
g_\tau = \mathrm{Encoder_g}(\{g^1_\tau, ..., g^i_\tau, ..., g^m_\tau\})
\end{equation}

Finally, a two-layer MLP integrates the multimodal information to obtain $C_\tau$. Performing the same operation on n game instructions in $I_G$ results in $C_G$.
\begin{equation}
C_\tau = \mathrm{MLP}(\mathtt{Concat}(d_\tau, f_\tau, g_\tau))
\end{equation}
\begin{equation}
C_G = \{C_1,..,C_\tau,...,C_n\}
\end{equation}

\subsubsection{Instruction Importance Estimation}
\label{importance}
In Section \ref{instruction}, we extract features from the instruction set $I_G = \{I_1\,..,I_\tau,...,I_n\}$ to derive the corresponding set of features $C_G = \{C_1,..,C_\tau,...,C_n\}$. However, the amount of contextual information regarding the task varies across instructions within $\{I_1\,..,I_\tau,...,I_n\}$. 
Our objective is to identify pivotal instructions and enhance their influence on DT.
Motivated by previous work~\citep{impt1, impt2}, we employ an attention mechanism to evaluate significance scores associated with each instruction in the set $I_G = \{I_1\,..,I_\tau,...,I_n\}$.
For a given instruction $I_\tau$, the importance score is calculated with:
\begin{equation}
S_i = \sum_{k=1}^{n} C_\tau \cdot C_k, \quad k \neq \tau
\end{equation}
By computing importance scores for each instruction, we obtain $S_G = \{S_1, \ldots, S_\tau, \ldots, S_n\}$, which are subsequently normalized using the softmax function:
\begin{equation}
S_\tau = \frac{e^{S_\tau}}{\sum_{k=1}^{n} e^{S_k}}, \quad k=1, \ldots, n
\end{equation}

\subsubsection{Instruction into DT}
\label{dt}
In Section \ref{instruction}, we obtain the features of Instruction set $I_G = \{I_1, \ldots, I_\tau, \ldots, I_n\}$: $C_G = \{C_1, \ldots, C_\tau, \ldots, C_n\}$ , along with their corresponding importance scores $S_G = \{S_1, \ldots, S_\tau, \ldots, S_n\}$ in Section \ref{importance}.
This section is dedicated to the integration of instructions into the DT.
We propose a novel design named \emph{SHyperGenerator} to infuse the information of $\{I_1, \ldots, I_\tau, \ldots, I_n\}$ and integrate parameters containing multimodal game instructions into the DT.
It possesses the capability to generate module parameters corresponding to game instructions and assigns weights to these parameters based on the importance scores $S_G = \{S_1, \ldots, S_\tau, \ldots, S_n\}$ associated with the instructions.
These modules, which represent diverse multimodal information, are affixed to the DT, thereby facilitating effective guidance.

The key idea of \emph{SHyperGenerator} is to use the representation of game instruction set $C_G = \{C_1, \ldots, C_\tau, \ldots, C_n\}$ as the input to the hypernetwork $\mathbf{H_{\theta}}$, generating a series of corresponding candidate parameters $\mathsf{P}=\{(\mathsf{D}_{1},\mathsf{U}_{1}),\cdots,(\mathsf{D}_n,\mathsf{U}_n)\}$ for Adapter of DT. 
At the same time, we utilize the importance scores $S_G = \{S_1, \ldots, S_\tau, \ldots, S_n\}$ as weights for $\mathsf{P}$, summing them up to obtain the information modules $\hat{\mathsf{P}} = {\textstyle \sum_{\tau =1}^{n}} \mathsf{P}_{\tau}S_{\tau}$ that are suitable for the current context:
\begin{equation}
    \mathsf{D}_\tau=\mathbf{H_\theta^{down}}({C}_\tau)=\mathsf{W_U^{down}}(\mathrm{Relu}(\mathsf{W_D^{down}}(C_\tau))
\end{equation}
\begin{equation}
    \mathsf{U}_\tau=\mathbf{H_\theta^{up}}({C}_\tau)=\mathsf{W_U^{up}}(\mathrm{Relu}(\mathsf{W_D^{up}}(C_\tau))
\end{equation}
\begin{equation}
    (\hat{\mathsf{D}},\hat{\mathsf{U}})={\textstyle \sum_{\tau=1}^{n}} S_{\tau}(\mathsf{D}_{\tau},\mathsf{U}_{\tau})
\end{equation}
where the hypernetworks $\mathbf{H_{\theta}^{down/up}}$ are used to generate the down-projection matrices $\mathsf{D}_{\tau}\in \mathbb{R} ^{d\times b}$ and the up-projection matrices $\mathsf{U}_{\tau}\in \mathbb{R} ^{b\times d}$ in the adapter. In particular, to limit the number of parameters, the hypernetworks $\mathbf{H_{\theta}^{down/up}}$ are designed using a bottleneck architecture: $\mathsf{W_D^{down/up}}\in \mathbb{R} ^{d'\times h}$ and $\mathsf{W_U^{down/up}}\in \mathbb{R} ^{h\times b\times d}$ are down-projection and up-projection matrices. $d$ is the input dimension of DT, $d'$ is the input dimension of encoded instruction, $b$ and $h$ are the bottleneck dimensions for the adapter layer and the hypernetwork, respectively.

Inspired by \citep{adapter_fusion,he2022towards}, we only insert these conditional parameters (Adapter) into the Feed-Forward Networks
(FFN) sub-layer in parallel:
\begin{equation}
    \mathbf{y}=\mathrm{FFN}(\mathrm{LN}(\mathbf{x}))+\mathrm{Adapter}(\mathrm{LN}(\mathbf{x}))
\end{equation}
where $\mathrm{LN}(\cdot)$ represents the LayerNorm layer, $\mathbf{y}$ is the output for the layer, and $\mathbf{x}$ is the output from the attention modules.

At this point, multimodal instruction is incorporated into the DT. 
Next, The DT takes the historical trajectory as the input and produces current actions as the output.

\section{Experiment}
\subsection{Setup}
\label{setup}
We select Atari ~\citep{atari} as our experimental environment for various reasons. 
Firstly, the visual observations in Atari games consist of pixels, enabling human descriptions of the game screens.
Secondly, the action space in Atari is discrete, making it easier to clearly distinguish the agent's current action. 
Lastly, Atari offers a diverse range of games with varying distributions, which facilitates the stimulation of the OOD ability of DT. 
We train our model and the baseline models using a subset of the samples in the DQN-replay dataset, observed by an online agent ~\citep{dqn}.
We evaluate ID ability in 37 training games and 10 unseen games, as reported in Appendix \ref{append_1}.
we report the mean and std of 3 seeds.
All raw scores are provided in Appendix \ref{append_2}.
Furthermore, to thoroughly assess the overall performance in multiple games, we standardize the scores. Specifically, we assign the highest positive score as 1 and the lowest negative score as -1. Subsequently, we linearly normalize the scores of all methods to the range of -1 to 1. 

To provide a clearer explanation of the objectives of our experiments, we delineate the individual setting for each experiment.
\begin{itemize}
    \item In Section \ref{exp_1}, we sample 10k transitions from the offline dataset for each of 37 training games, amounting to a total of approximately 400k transitions utilized for training both our model and the baseline models. \textbf{This section investigates whether multimodal game instruction enhances the multitasking and generalization of the decision transformer.}
    \item In Section \ref{exp_2}, we sample 10k, 100k, and 200k transitions from the offline dataset for each of 37 training games, amounting to a total of approximately 400k, 4M, and 8M transitions for training both our model and the DT model. \textbf{This section explores the impact of the training dataset size on the model's performance.}
    \item In Section \ref{exp_3}, we sample 10k transitions from the offline dataset for each of 10 games, 20 games, and 37 games, amounting to a total of approximately 100k, 200k, and 400k transitions for training our model. \textbf{This section investigates whether the number of training games influences the OOD capabilities of the model.}
    \item Experimental setting of \ref{exp_4} is the same as \ref{exp_1}. \textbf{We use DTGI-a(assign equal scores to each instruction) to explore whether the model focuses on critical aspects of game instruction.}
\end{itemize}

\subsection{Baselines}
We compare our proposed DTGI with four baselines.

\textbf{1. Decision Transformer (DT):} DT is trained to learn multiple tasks from the training set without task information. The training process is consistent with our method. This baseline helps assess the impact of task context.\\
\textbf{2. Decision Transformer with Textual Language (DTL):} In this baseline, we provide a text description for each game, serving as contextual information. To ensure fairness, the only difference between our method and DTL is the inclusion of context information. The model structure and training process remain unchanged. DTL is utilized to demonstrate that depending solely on a single text as task contextual information is inadequate.\\
\textbf{3. Decision Transformer with Vision Trajectory (DTV):} Here, we provide an expert trajectory for each game as contextual information. To maintain fairness, the model structure and training process are identical to our method. DTV illustrates that a single trajectory is insufficient in providing enough task context.\\
\textbf{4. Decision Transformer with Game Instruction - average (DTGI-a):} This baseline diverges from DTGI by assigning an equal score to every instruction within the game instruction set. DTGI-a is designed to ablate Instruction Importance Estimation in Section \ref{importance}.

\subsection{Does multimodal game instruction enhance the multitasking and generalization abilities of the decision transformer?}
\label{exp_1}
In this section, we compare the ID and OOD ability of DTGI and the baselines. Our investigation seeks to determine if Multimodal Game Instruction (MGI) facilitates multitasking and enhances generalization ability. Additionally, we examine whether MGI provides sufficient task context information and assess its effectiveness compared to text and trajectory. To evaluate multitasking capabilities, we examine the performance scores across 37 training games from the set $S_{\text{train}}$. The ID results are presented in Table \ref{table_id}.
Concurrently, we evaluate generalization abilities by analyzing scores from 10 unseen games in the set  $S_{\text{test}}$. OOD results are displayed in Table \ref{table_ood}.

We observe that: 
(1) The integration of contextual information, such as textual language, visual trajectory, and multimodal instruction, significantly enhances the multitasking and generalization capabilities of DT. Incorporating contextual information notably enhances the effective facilitation of a universal network in performing multiple tasks.
(2) Multimodal instruction surpasses both textual language and visual trajectory, underscoring its ability to offer more detailed and comprehensive task context information. It is imperative to underscore that DTGI demonstrates significant advantages in handling unseen games.

\subsection{Does the number of training games influence the model's OOD capabilities?}
\label{exp_3}
In this section, we investigate the impact of the quantity of training games on the OOD performance of DTGI. Augmenting the number of training games not only enriches data diversity but also exposes the model to a broader spectrum of instructions, fostering knowledge sharing among them. DTGI undergoes training with three different configurations: 10 games (25\%), 20 games (50\%), and the entirety of 37 games (100\%). The OOD results are elucidated in Table \ref{table_5}.

We observe that with an increase in the number of trained games, the OOD performance of the model demonstrates significant improvement. 
OOD performance exhibits a pronounced sensitivity to the quantity of training instances. 
Incorporating diverse gaming tasks in the training process is advisable to improve the model's OOD performance comprehensively. 
This approach fosters the exchange of knowledge among different tasks.

\begin{table}
\begin{minipage}[!ht]{0.48\textwidth}
\centering
\resizebox{1.00\linewidth}{!}{
\begin{tabular}{c|ccccc}
\toprule
\textbf{G} & \textbf{DT}  & \textbf{DTL}  & \textbf{DTV}  & \textbf{DTGI-a} & \textbf{DTGI}  \\ \midrule
1      & -0.96 \tiny{$\pm$ 0.22}         & -0.91 \tiny{$\pm$ 0.17}            & -0.88 \tiny{$\pm$ 0.09}            & -1.00 \tiny{$\pm$ 0.18}    & \textbf{-0.63 \tiny{$\pm$ 0.14}}  \\
2      & 0.69 \tiny{$\pm$ 0.09}          & 0.71 \tiny{$\pm$ 0.04}            & 0.73 \tiny{$\pm$ 0.03}           & \textbf{1.00 \tiny{$\pm$ 0.06}}           & 0.92 \tiny{$\pm$ 0.04}           \\
3      & 0.00 \tiny{$\pm$ 0.00}          & 0.00 \tiny{$\pm$ 0.00}            & 0.00 \tiny{$\pm$ 0.00}           & 0.00 \tiny{$\pm$ 0.00}           & 0.00 \tiny{$\pm$ 0.00} \\
4      & 0.51 \tiny{$\pm$ 0.10}          & 0.62 \tiny{$\pm$ 0.00}           & 0.64 \tiny{$\pm$ 0.05}            & \textbf{1.00 \tiny{$\pm$ 0.04}}           & 0.66 \tiny{$\pm$ 0.03}         \\
5      & 0.00 \tiny{$\pm$ 0.00}          & 0.00 \tiny{$\pm$ 0.00}           & 0.00 \tiny{$\pm$ 0.00}            & 0.00 \tiny{$\pm$ 0.00}           & \textbf{1.00 \tiny{$\pm$ 0.70}}         \\
6      & -0.79 \tiny{$\pm$ 0.06}          & -0.70 \tiny{$\pm$ 0.04}           & \textbf{-0.55 \tiny{$\pm$ 0.12}}           & -1.00 \tiny{$\pm$ 0.02}           & -0.72 \tiny{$\pm$ 0.02} \\
7      & 0.51 \tiny{$\pm$ 0.09}          & 0.64 \tiny{$\pm$ 0.18}           & 0.21 \tiny{$\pm$ 0.02}           & 0.81 \tiny{$\pm$ 0.12}         & \textbf{1.00 \tiny{$\pm$ 0.37}}        \\ 
8      & 0.12 \tiny{$\pm$ 0.04}         & 0.29 \tiny{$\pm$ 0.08}            & 0.53 \tiny{$\pm$ 0.26}          & 0.88 \tiny{$\pm$ 0.19}        & \textbf{1.00 \tiny{$\pm$ 0.21}}               \\                
9      & 0.80 \tiny{$\pm$ 0.07}         & 0.80 \tiny{$\pm$ 0.31}         & 0.50 \tiny{$\pm$ 0.07}          & 0.90 \tiny{$\pm$ 0.12}          & \textbf{1.00 \tiny{$\pm$ 0.07}}           \\                     
10     & 0.60 \tiny{$\pm$ 0.13}         & 0.60 \tiny{$\pm$ 0.08}          & 0.80 \tiny{$\pm$ 0.15}        & 0.73 \tiny{$\pm$ 0.06}          & \textbf{1.00 \tiny{$\pm$ 0.16}}                \\
11     & 0.56 \tiny{$\pm$ 0.07}         & 0.82 \tiny{$\pm$ 0.00}          & 0.60 \tiny{$\pm$ 0.04}         & \textbf{1.00 \tiny{$\pm$ 0.02}}          & 0.73 \tiny{$\pm$ 0.03}                \\
12     & 0.56 \tiny{$\pm$ 0.12}          & 0.74 \tiny{$\pm$ 0.15}          & 0.77 \tiny{$\pm$ 0.17}          & \textbf{1.00 \tiny{$\pm$ 0.20}}         & 0.76 \tiny{$\pm$ 0.14}                 \\
13     & 0.53 \tiny{$\pm$ 0.09}          & 0.13 \tiny{$\pm$ 0.09}           & 0.60 \tiny{$\pm$ 0.08}          & \textbf{1.00 \tiny{$\pm$ 0.08}}          & 0.80 \tiny{$\pm$ 0.24}               \\
14     & 0.88 \tiny{$\pm$ 0.24}          & 0.25 \tiny{$\pm$ 0.05}           & 0.52 \tiny{$\pm$ 0.05}          & 0.27 \tiny{$\pm$ 0.07}          & \textbf{1.00 \tiny{$\pm$ 0.40}}     \\
15     & 0.47 \tiny{$\pm$ 0.08}         & \textbf{1.00 \tiny{$\pm$ 0.04}}        & 0.03 \tiny{$\pm$ 0.02}          & \textbf{1.00 \tiny{$\pm$ 0.19}}          & 0.73 \tiny{$\pm$ 0.24}             \\          
16     & \textbf{-0.26 \tiny{$\pm$ 0.05}}         & -0.52 \tiny{$\pm$ 0.09}          & -0.43 \tiny{$\pm$ 0.03}           & \textbf{-0.26 \tiny{$\pm$ 0.09}}          & -1.00 \tiny{$\pm$ 0.06}         \\
17     & 0.08 \tiny{$\pm$ 0.03}         & \textbf{1.00 \tiny{$\pm$ 0.05}}          & 0.70 \tiny{$\pm$ 0.01}          & 0.49 \tiny{$\pm$ 0.01}          & 0.98 \tiny{$\pm$ 0.04}               \\
18     & 0.74 \tiny{$\pm$ 0.09}       & \textbf{1.00 \tiny{$\pm$ 0.17}}          & 0.72 \tiny{$\pm$ 0.29}          & 0.68 \tiny{$\pm$ 0.07}           & 0.83 \tiny{$\pm$ 0.16}                  \\
19     & 0.00 \tiny{$\pm$ 0.00}         & 0.00 \tiny{$\pm$ 0.00}          & 0.00 \tiny{$\pm$ 0.00}         & 0.00 \tiny{$\pm$ 0.00}          & 0.00 \tiny{$\pm$ 0.00}                     \\
20     & 0.00 \tiny{$\pm$ 0.00}         & 0.00 \tiny{$\pm$ 0.00}         & 0.00 \tiny{$\pm$ 0.00}         & 0.00 \tiny{$\pm$ 0.00}          & 0.00 \tiny{$\pm$ 0.00}                     \\
21     & 0.41 \tiny{$\pm$ 0.11}        & 0.47 \tiny{$\pm$ 0.11}          & 0.94 \tiny{$\pm$ 0.18}          & \textbf{1.00 \tiny{$\pm$ 0.33}}           & 0.59 \tiny{$\pm$ 0.17}             \\
22     & 0.88 \tiny{$\pm$ 0.20}        & 0.11 \tiny{$\pm$ 0.03}         & 0.49 \tiny{$\pm$ 0.20}           & \textbf{1.00 \tiny{$\pm$ 0.16}}           & 0.88 \tiny{$\pm$ 0.14}                  \\
23     & 0.34 \tiny{$\pm$ 0.24}          & \textbf{1.00 \tiny{$\pm$ 0.30}}        & 0.93 \tiny{$\pm$ 0.29}          & 0.71 \tiny{$\pm$ 0.16}          & 0.81 \tiny{$\pm$ 0.27}               \\
24     & \textbf{1.00 \tiny{$\pm$ 00.17}}          & \textbf{1.00 \tiny{$\pm$ 0.17}}          & 0.82 \tiny{$\pm$ 0.11}          & 0.45 \tiny{$\pm$ 0.23}          & 0.55 \tiny{$\pm$ 0.00}            \\
25     & 0.75 \tiny{$\pm$ 0.06}          & 0.96 \tiny{$\pm$ 0.15}          & 0.51 \tiny{$\pm$ 0.02}         & \textbf{1.00 \tiny{$\pm$ 0.26}}          & 0.53 \tiny{$\pm$ 0.11}                \\
26     & 0.16 \tiny{$\pm$ 0.02}         & 0.45 \tiny{$\pm$ 0.18}           & \textbf{1.00 \tiny{$\pm$ 0.01}}          & 0.75 \tiny{$\pm$ 0.18}         & 0.20 \tiny{$\pm$ 0.01}                     \\
27     & 0.87 \tiny{$\pm$ 0.08}         & \textbf{1.00 \tiny{$\pm$ 0.13}}          & 0.65 \tiny{$\pm$ 0.09}          & 0.58 \tiny{$\pm$ 0.08}          & 0.58 \tiny{$\pm$ 0.14}              \\
28     & 0.05 \tiny{$\pm$ 0.00}         & 0.23 \tiny{$\pm$ 0.03}          & 0.40 \tiny{$\pm$ 0.05}        & 0.15 \tiny{$\pm$ 0.05}        & \textbf{1.00 \tiny{$\pm$ 0.19}}                   \\
29     & \textbf{1.00 \tiny{$\pm$ 0.22}}         & 0.70 \tiny{$\pm$ 0.05}          & 0.96 \tiny{$\pm$ 0.23}         & 0.81 \tiny{$\pm$ 0.00}          & 0.92 \tiny{$\pm$ 0.12}        \\
30     & 0.20 \tiny{$\pm$ 0.03}         & 0.64 \tiny{$\pm$ 0.13}          & \textbf{1.00 \tiny{$\pm$ 0.17}}          & 0.57 \tiny{$\pm$ 0.09}          & 0.89 \tiny{$\pm$ 0.10}               \\
31     & 0.19 \tiny{$\pm$ 0.13}         & 0.47 \tiny{$\pm$ 0.03}          & 0.78 \tiny{$\pm$ 0.35}          & 0.74 \tiny{$\pm$ 0.21}         & \textbf{1.00 \tiny{$\pm$ 0.32}}           \\
32     & 0.69 \tiny{$\pm$ 0.08}          & 0.92 \tiny{$\pm$ 0.05}          & 0.94 \tiny{$\pm$ 0.23}          & 0.65 \tiny{$\pm$ 0.05}         & \textbf{1.00 \tiny{$\pm$ 0.12}}                \\
33     & 0.76 \tiny{$\pm$ 0.08}        & 0.74 \tiny{$\pm$ 0.14}          & 0.71 \tiny{$\pm$ 0.13}          & \textbf{1.00 \tiny{$\pm$ 0.25}}          & 0.85 \tiny{$\pm$ 0.19}                    \\
34     & -0.72 \tiny{$\pm$ 0.00}         & -0.72 \tiny{$\pm$ 0.00}          & -0.72 \tiny{$\pm$ 0.00}          & -1.00 \tiny{$\pm$ 0.00}          & -0.72 \tiny{$\pm$ 0.00}                     \\
35     & 0.28 \tiny{$\pm$ 0.04}         & 0.73 \tiny{$\pm$ 0.13}          & 0.60 \tiny{$\pm$ 0.15}          & \textbf{1.00 \tiny{$\pm$ 0.20}}          & 0.43 \tiny{$\pm$ 0.11}                     \\
36     & 0.67 \tiny{$\pm$ 0.08}         & 0.57 \tiny{$\pm$ 0.07}           & \textbf{1.00 \tiny{$\pm$ 0.10}}          & 0.58 \tiny{$\pm$ 0.10}           & 0.65 \tiny{$\pm$ 0.04}                    \\
37     & 0.00 \tiny{$\pm$ 0.00}         & 0.31 \tiny{$\pm$ 0.11}           & 0.77 \tiny{$\pm$ 0.20}          & \textbf{1.00 \tiny{$\pm$ 0.33}}          & 0.98 \tiny{$\pm$ 0.09}                    \\
\rowcolor[HTML]{EFEFEF} 
\textbf{O}            & 0.42 \tiny{$\pm$ 0.08}         & 0.51 \tiny{$\pm$ 0.09}          & 0.54 \tiny{$\pm$ 0.10}         & 0.61 \tiny{$\pm$ 0.11}         & \textbf{0.66 \tiny{$\pm$ 0.12}}   \\ 
\bottomrule
\end{tabular}
}
\vspace{2pt}
\caption{ID results (37 training games)}
\label{table_id}
\end{minipage}
\hfill
\begin{minipage}[!htbp]{0.48\textwidth}
\centering
\resizebox{1.00\linewidth}{!}{
\begin{tabular}{c|ccccc}
\toprule
\textbf{G} & \textbf{DT}  & \textbf{DTL}  & \textbf{DTV}  & \textbf{DTGI-a} & \textbf{DTGI}   \\ 
\midrule
1  &  -1.00 \tiny{$\pm$ 0.22}           &  -0.48 \tiny{$\pm$ 0.23}           & -0.38 \tiny{$\pm$ 0.14}           & \textbf{1.00 \tiny{$\pm$ 0.00}}           & -0.70 \tiny{$\pm$ 0.27} \\
2       &  0.00 \tiny{$\pm$ 0.00}           & 0.20 \tiny{$\pm$ 0.14}           & 0.80 \tiny{$\pm$ 0.28}           & \textbf{1.00 \tiny{$\pm$ 0.71}}           & 0.80 \tiny{$\pm$ 0.28}                  \\ 
3   & -0.36 \tiny{$\pm$ 0.18}          & -0.64 \tiny{$\pm$ 0.00}           & -0.82 \tiny{$\pm$ 0.14}           & -1.00 \tiny{$\pm$ 0.14}           & \textbf{-0.21 \tiny{$\pm$ 0.08}}      \\
4                 &  0.00 \tiny{$\pm$ 0.00}         & 0.50 \tiny{$\pm$ 0.35}           & 0.00 \tiny{$\pm$ 0.00}          & 0.00 \tiny{$\pm$ 0.00}          & \textbf{1.00 \tiny{$\pm$ 0.35}}      \\ 
5                 &  \textbf{1.00 \tiny{$\pm$ 0.13}}         & 0.31 \tiny{$\pm$ 0.05}          & 0.10 \tiny{$\pm$ 0.03}          & 0.33 \tiny{$\pm$ 0.02}          & 0.52 \tiny{$\pm$ 0.03}      \\ 
6           &  \textbf{1.00 \tiny{$\pm$ 0.12}}         & 0.96 \tiny{$\pm$ 0.14}        & 0.73 \tiny{$\pm$ 0.12}          & 0.96 \tiny{$\pm$ 0.12}           & 0.73 \tiny{$\pm$ 0.12}                  \\
7          & 0.50 \tiny{$\pm$ 0.06}         & 0.40 \tiny{$\pm$ 0.07}         & \textbf{1.00 \tiny{$\pm$ 0.13}}          & 0.40 \tiny{$\pm$ 0.28}          & 0.20 \tiny{$\pm$ 0.08}             \\
8          & 0.28 \tiny{$\pm$ 0.08}         & \textbf{1.00 \tiny{$\pm$ 0.09}}         & 0.11 \tiny{$\pm$ 0.06} 
& 0.19 \tiny{$\pm$ 0.04}          & 0.61 \tiny{$\pm$ 0.11}             \\
9                &  0.00 \tiny{$\pm$ 0.00}         & 0.83 \tiny{$\pm$ 0.36}          & \textbf{1.00 \tiny{$\pm$ 0.00}}          & 0.83 \tiny{$\pm$ 0.12}           & 0.92 \tiny{$\pm$ 0.16}           \\
10         &  0.93 \tiny{$\pm$ 0.02}         & 0.50 \tiny{$\pm$ 0.15}         & 0.45 \tiny{$\pm$ 0.21}           & 0.84 \tiny{$\pm$ 0.06}          & \textbf{1.00 \tiny{$\pm$ 0.30}}                     \\
\rowcolor[HTML]{EFEFEF}
\textbf{O}            & 0.24 \tiny{$\pm$ 0.08}         & 0.36 \tiny{$\pm$ 0.15}          & 0.30 \tiny{$\pm$ 0.11}          & 0.46 \tiny{$\pm$ 0.15}          & \textbf{0.49 \tiny{$\pm$ 0.18}}                    \\ \bottomrule
\end{tabular}
}
\vspace{2pt}
\caption{OOD results (10 unseen games)}
\label{table_ood}
\vspace{1.0cm}
\centering
\resizebox{1.00\linewidth}{!}{
\begin{tabular}{l|lllll}
\toprule
\textbf{G} & \textbf{10 Games(25\%)}  & \textbf{20 Games(50\%)}   & \textbf{37 Games(100\%)}    \\ 
\midrule
1  &  \textbf{1.00  \tiny{$\pm$ 0.04}}          & -0.80  \tiny{$\pm$ 0.4}           & -1.00  \tiny{$\pm$ 0.39}         \\
2       &  0.00  \tiny{$\pm$ 0.00}           & 0.00  \tiny{$\pm$ 0.00}           & \textbf{1.00  \tiny{$\pm$ 0.35}}  \\ 
3   & -1.00  \tiny{$\pm$ 0.00}          & -1.00  \tiny{$\pm$ 0.22}           & \textbf{-0.40  \tiny{$\pm$ 0.16}}  \\
4                 &  0.00  \tiny{$\pm$ 0.00}         & \textbf{1.00  \tiny{$\pm$ 0.70}}           & 0.10  \tiny{$\pm$ 0.04} \\ 
5                 &  0.80  \tiny{$\pm$ 0.06}         & 0.84  \tiny{$\pm$ 0.08}           & \textbf{1.00  \tiny{$\pm$ 0.06}}        \\ 
6           &  0.36  \tiny{$\pm$ 0.00 }        & 0.82  \tiny{$\pm$ 0.20}        & \textbf{1.00  \tiny{$\pm$ 0.17}}                          \\
7         & \textbf{1.00  \tiny{$\pm$ 0.24}}         & 0.54  \tiny{$\pm$ 0.05}         & 0.15  \tiny{$\pm$ 0.06}        \\
8          & 0.05  \tiny{$\pm$ 0.02}         & 0.00  \tiny{$\pm$ 0.00}         & \textbf{1.00  \tiny{$\pm$ 0.18}}
           \\
9                &  0.09  \tiny{$\pm$ 0.06}         & 0.64  \tiny{$\pm$ 0.06}          & \textbf{1.00  \tiny{$\pm$ 0.17}}                        \\
10         &  0.17  \tiny{$\pm$ 0.03}         & \textbf{1.00  \tiny{$\pm$ 0.10}}         & 0.82  \tiny{$\pm$ 0.25}                                \\
\rowcolor[HTML]{EFEFEF}
\textbf{O}            & 0.25  \tiny{$\pm$ 0.05}         & 0.30  \tiny{$\pm$ 0.18}           & \textbf{0.47  \tiny{$\pm$ 0.18}}                 \\ 
\bottomrule
\end{tabular}
}
\vspace{2pt}
\caption{OOD results of DTGI are acquired using offline datasets comprising 10, 20, and 37 games.}
\label{table_5}
\end{minipage}
\end{table}

\subsection{Does the size of the training dataset affect the model's performance?}
\label{exp_2}
In this section, We consider whether the size of the dataset has an impact on model performance. 
As the dataset expands, concomitant with increased in-distribution data diversity, there is a consistent enhancement in performance.
we sample 10k, 100k, and 200k transitions from the offline dataset for each of the 37 training games, employing them to train both DT and our model. 
The overall scores of ID evaluation and OOD evaluation across multiple games are illustrated in Figure \ref{fig_exp}, and individual game scores are detailed in Appendix \ref{append_2}.

We observe that: 
(1) The dataset expansion contributes to enhanced performance in both DT and our model. Notably, our model consistently outperforms DT.
(2) In contrast to OOD performance, ID performance shows sensitivity to changes in dataset size. By increasing the dataset size, it is difficult to greatly improve the OOD ability of the model.
(3) In the ID setting, our model consistently enhances performance with an increase in dataset size. However, the performance of DT reaches a saturation point when the dataset expands from 100k to 200k samples.
\begin{figure}[!htbp]
\centering
\includegraphics[width=0.55\linewidth]{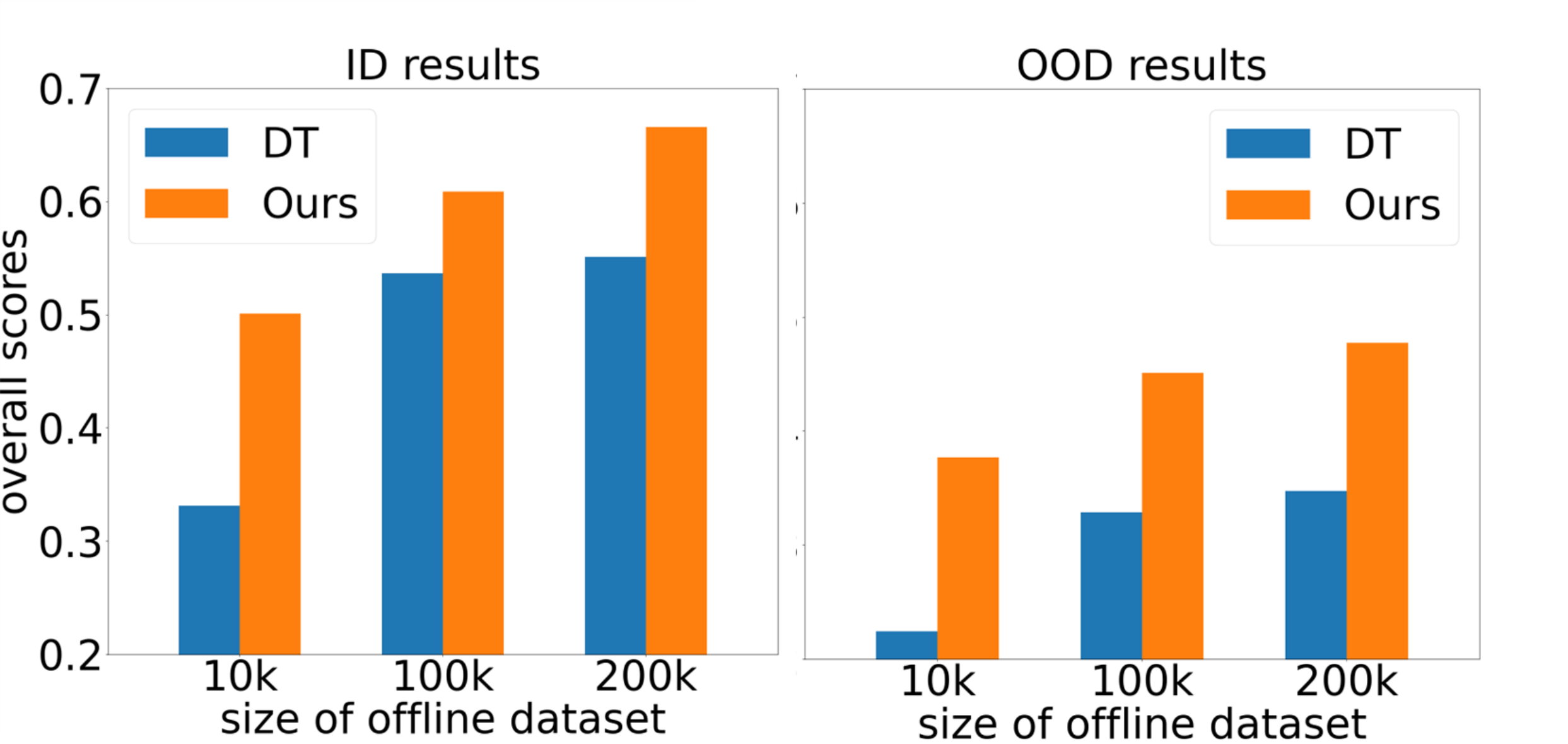}
\caption{Performance comparison of DT and our model under different dataset sizes.} 
\label{fig_exp}
\end{figure}

\subsection{Does DTGI focus on the critical parts of game instruction?}
\label{exp_4}
\begin{figure}[!htbp]
\centering
\includegraphics[width=0.60\linewidth]{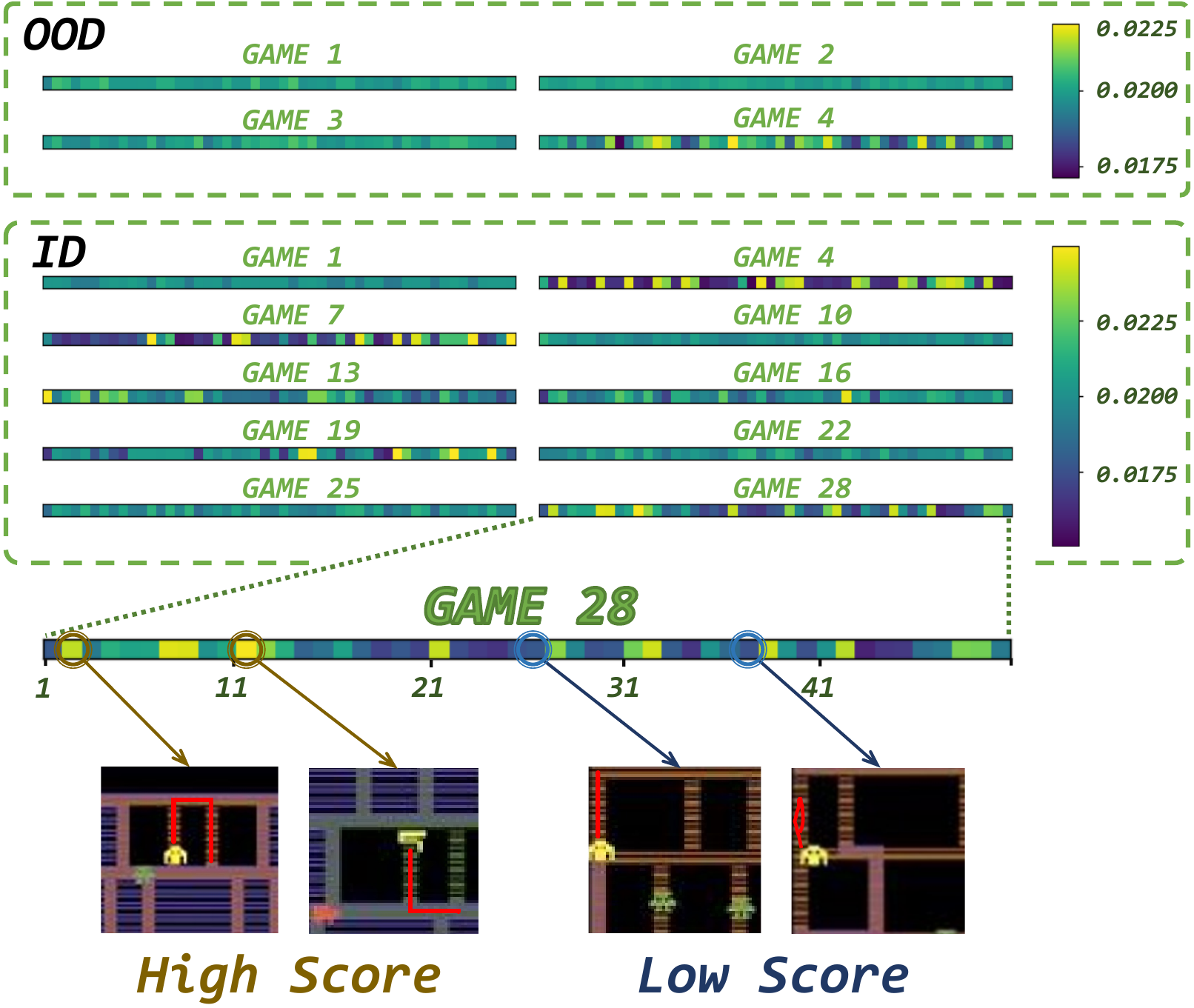}
\caption{Visualization of Instruction Importance scores for 10 training games and 4 unseen games and an in-depth analysis of the 28th training game reveals a correlation between higher scores and increased trajectory diversity.}
\label{fig_v}
\end{figure}

In this section, we investigate the enhancement of Instruction Importance Estimation (see Section \ref{importance}) by assigning importance scores to multiple instructions.
In Table \ref{table_id} and \ref{table_ood}, we list the results of ID results and OOD results of DTGI and DTGI-a(assign equal scores to each instruction). 
The instruction set of each game contains 50 instructions. 
The importance scores for these instructions, visualized in Figure \ref{fig_v}, are selectively shown for a subset of games, while comprehensive results for all games are available in Appendix \ref{append_4}. 
The analysis specifically highlights the 28th training game, where the score of DTGI is 1.00 and the score of DTGI-a is 0.15.

we observe that Instruction Importance Estimation enhances model performance, as evidenced by the data in Tables \ref{table_id} and \ref{table_ood}.
An in-depth examination of the 28th training game demonstrates the correlation between elevated instruction scores and a broader diversity in the trajectory of multimodal game instruction.

\section{Related Work}
\subsection{Task Conditioned Policy}
\begin{table*}[!htbp]
\small
\centering
\begin{tabular}{lllll}
\toprule[0.20\heavyrulewidth] 
\textbf{Method}         & \textbf{Condition Type}    & \textbf{ID evaluation}   & \textbf{OOD evaluation}   & \textbf{Environment} \\ \midrule
\citet{chessgpt}        & Text                       & \checkmark  & $\times$ & Chess  \\
\citet{reward23}        & Text                       & \checkmark  & \checkmark & Robot  \\
\citet{concept23}       & Text                       & \checkmark & $\times$ & RTFM \\
\citet{controller23}    & Text                       & \checkmark &  $\times$ & Minecraft \\
\citet{emach23}         & Text                       & \checkmark & $\times$ &  MiniGrid \\
\citet{planner22}        & Text                       & \checkmark & $\times$ & Kitchen \\
\citet{deps23}         & Text                       & \checkmark & $\times$ &  Minecraft \\
\midrule
\citet{hyperdt23}       & Trajectory                 & $\times$ & \checkmark & Meta World  \\
\citet{groot23}        & Trajectory                 &  \checkmark & $\times$ & Minecraft \\
\citet{icdt23}          & Trajectory                 & $\times$  & \checkmark & MiniHack \\
\midrule
R2-Play (Ours)          & Multimodal                 & \checkmark & \checkmark & Atari \\ \bottomrule
\end{tabular}
\caption{Comparisons between ours and related works regarding condition type, ID/OOD evaluation, and environment.}
\label{table_related}
\end{table*}

Task-conditioned policy, represented as $\mathbf{\pi}(a | s, t)$, signifies a policy that is conditioned on a particular task. 
This policy takes as input the current state ($s$) and task information ($t$), producing an action ($a$) as output. 
The integration of task information empowers a unified policy to effectively handle diverse tasks.
Recently, numerous works utilize texts as task conditions.
For instance, ~\citet{reward22, minedojo22, reward23, chessgpt} calculate the similarity score between textual tasks and visual observations, utilizing this score as a reward function to train a multi-task RL agent.
~\citet{concept23, controller23, emach23} concentrate on acquiring visual state representations relevant to the text, facilitating improved integration into the decision network.
When addressing demanding long-horizon tasks, ~\citet{planner22, deps23, auto} utilize LLMs to generate step-by-step high-level plans guiding the agents.
Additionally, ~\citet{hyperdt23, icdt23, groot23} employ trajectories as task conditions.
While both textual language and visual trajectory offer advantages, they also present distinct limitations. Text is not expressive enough for visual-based decision tasks while extracting a correct strategy from trajectory proves challenging without contextual task information.
To overcome these limits, we construct a set of Multimodal Game Instructions to provide rich and detailed context.
We compare ``Read to Play'' with related works across multiple dimensions in Table~\ref{table_related}. 

\subsection{Multimodal Instruction Tuning}
In the field of NLP, recent works underscore the significance of instruction tuning~\citep{instructsurvey, inlp} as a crucial technique to enhance the capabilities and controllability of LLMs, such as GPT-3~\citep{gpt3} and InstrctGPT~\citep{instructgpt}.
Instruction tuning allows these models to effectively follow instructions and execute new commands using only a few in-context learning examples. 
Recently, similar efforts have been made in the multimodal field.
Multi-Instruct~\citep{1mit23} initially introduces instruction tuning, organizing 47 diverse multimodal tasks across 11 categories.
LLaMA-Adapter~\citep{llama-adapter23} further extends this approach by incorporating additional adapter modules and multimodal prompts, thus adapting LLaMA into an instruction-following model.
LLavA~\citep{vi23} provides a pipeline to convert image-text pairs into instruction-following data using GPT.
Some works~\citep{otter23, mmrlhf23, mmic23} focus on training a multimodal model with in-context instruction tuning, without embedding visual information into the language model, and achieve impressive results. 
However, in the field of RL, the application of instruction tuning remains unexplored. 
If vision-based RL tasks are perceived as long-horizon vision tasks, introducing multimodal instruction tuning to RL is promising. 
Taking inspiration from these works on multimodal instruction, we construct a set of multimodal game instructions for decision control.

\subsection{Hypernet Adapter}
The method of Adapter tuning~\citep{pmlr-v97-houlsby19a} is initially developed in the field of NLP.
It is a commonly used and parameter-efficient approach to fine-tuning pre-trained language models~\citep{peft1, peft2, peft3, peft4, peft5, peft6}.
The main idea is to train a compact module, referred to as an Adapter, which can subsequently be adapted for downstream tasks.
However, in multi-task learning, this method requires learning different modules for each task, increasing parameter costs as the number of tasks increases. 
In contrast, task-specific modules are independent and do not facilitate knowledge transfer between them. 
Recent works~\citep{karimi-mahabadi-etal-2021-parameter, ivison-peters-2022-hyperdecoders, zhao-etal-2023-prototype} propose training a hypernetwork to generate parameters for these modules, thus achieving an optimal balance between parameter efficiency and adaptation for downstream tasks.
These methods encourage the multi-task learning model to capture shared information by utilizing a task-shared hypernetwork, while also avoiding negative task interference by generating conditioned modules individually.
However, these methods typically use coarse-grained task embeddings as contexts for generating parameters, which cannot precisely guide agents in the RL setting. 
In our work, we propose using fine-grained multimodal instruction as the context for generating parameters with a hypernetwork, which can better guide the decision-making of the agent.

\section{Conclusion}
In this paper, we construct a comprehensive multimodal game instruction set, offering extensive context for a variety of games.
The experimental findings reveal that integrating multimodal instructions significantly improves performance, outperforming the results obtained with text or trajectory guidance.
Our analysis suggests that broadening the number of games in the dataset enhances the model's OOD performance more effectively than merely increasing the dataset size.

In the field of LLMs, instruction tuning emerges as a pivotal technology for augmenting the generalization capabilities of models.
Inspired by advancements in multimodal instruction tuning ~\citep{1mit23, llama-adapter23, otter23, vi23}, this paper presents an innovative application of this technique in the context of decision control.
To the best of our knowledge, this is the first attempt to integrate multimodal instruction tuning into RL.
Future research should investigate the feasibility of a generalist multimodal instruction framework, aiming to enhance performance in both vision tasks and vision-based RL tasks.

\bibliographystyle{apalike}
\bibliography{main}

\clearpage
\appendix
\onecolumn
\section{Appendix}

\subsection{Training Games and Unseen Games}
\label{append_1}

\begin{table}[!htbp]
\centering
\resizebox{0.4\linewidth}{!}{
\begin{tabular}{l|ll|l}
\toprule
id & training game  & id   & unseen game    \\ 
\midrule
1          &  Journey\_Escape         & 1           & Private\_Eye         \\
2          &  Krull          & 2           & Breakout  \\ 
3          & Montezuma\_Revenge           & 3          & Ice\_Hockey  \\
4          &  Bowling         & 4           & Hero \\ 
5          &  Solaris         & 5           & Phoenix        \\ 
6          &  Fishing\_Derby         & 6       & Demon\_Attack                          \\
7          & Space\_Invaders         & 7        & Berzerk       \\
8          & Battle\_Zone         & 8        & Pooyan   \\
9          &  Kangaroo         & 9          & Frostbite                       \\
10         &  Seaquest         & 10         & Defender                             \\
11         &  Name\_This\_Game         &          &                               \\
12         &  Atlantis          &          &       \\
13         &  Jamesbond           &            &  \\ 
14         & Centipede           &            &  \\
15         &  QBert        &            &  \\ 
16         &  Double\_Dunk         &            &  \\ 
17         & Kung\_Fu\_Master         &         &                  \\
18         & Riverraid        &          &      \\
19         & Elevator\_Action         &         &  \\
20         &  Venture        &           &           \\
21         &  Asterix         &          &                               \\
22         &  Enduro         &          &                              \\
23         &  Road\_Runner          &            &         \\
24         &  Robotank           &           &   \\ 
25         & Carnival           &            &  \\
26         &  Time\_Pilot        &            & \\ 
27         &  Beam\_Rider        &            &        \\ 
28         &  Amidar         &         &                      \\
29         & Star\_Gunner         &          &         \\
30         & Up\_N\_Down         &          &        \\
31         &  Boxing         &          &                        \\
32         &  Yars\_Revenge        &          &                            \\
33         &  Alien         &          &                              \\
34         &  Skiing         &           &                    \\
35         &  Crazy\_Climber         &          &                              \\
36         &  Assault         &          &                               \\
37         &  Gopher         &          &                           \\ 
\bottomrule
\end{tabular}
}
\label{table_6}
\vspace{1ex}
\caption{Game list}
\end{table}

\newpage
\subsection{Raw Scores in the Experiment}
\label{append_2}
1. \textbf{Raw Scores in the Experiment 5.3}
\begin{table}[H]
\centering
\resizebox{0.88\linewidth}{!}{
\begin{tabular}{c|ccccc}
\toprule
\textbf{G} & \textbf{DT}  & \textbf{DTL}  & \textbf{DTV}  & \textbf{DTGI-a} & \textbf{DTGI}  \\ \midrule
1        &    -25900.0$\pm$5799.57    &    -24700.00$\pm$ 4677.79    &    -23833.33$\pm$2417.41    &    -26900.00$\pm$4899.66    &    -17066.67$\pm$3860.12    \\
2        &    2406.67$\pm$320.16    &    2486.67$\pm$138.88    &    2536.67$\pm$110.63    &    3490.00$\pm$196.51    &    3206.67$\pm$151.35    \\
3        &    0.00$\pm$0.00    &    0.00$\pm$0.00    &    0.00$\pm$0.00    &    0.00$\pm$0.00    &    0.00$\pm$0.00    \\
4        &    12.33$\pm$2.46    &    15.00$\pm$0.00    &    15.67$\pm$1.25    &    24.33$\pm$1.03    &    16.00$\pm$0.71    \\
5        &    0.00$\pm$0.00    &    0.00$\pm$0.00    &    0.00$\pm$0.00    &    0.00$\pm$0.00    &    486.67$\pm$344.13    \\
6        &    -64.00$\pm$4.97    &    -56.00$\pm$3.56    &    -44.67$\pm$10.08    &    -81.00$\pm$1.87    &    -58.00$\pm$1.63    \\
7        &    173.33$\pm$30.57    &    218.33$\pm$61.04    &    71.67$\pm$8.25    &    276.67$\pm$39.81    &    341.67$\pm$126.12    \\
8        &    666.67$\pm$235.7    &    1666.67$\pm$471.40   &    3000.00$\pm$1471.96    &    5000.00$\pm$1080.12    &    5666.67$\pm$1178.51    \\
9        &    533.33$\pm$47.14    &    533.33$\pm$205.48    &    333.33$\pm$47.14    &    600.00$\pm$81.65    &    666.67$\pm$47.14    \\
10        &    160.00$\pm$35.59    &    160.00$\pm$21.60   &    213.33$\pm$41.10    &    193.33$\pm$17.00    &    266.67$\pm$41.90    \\
11        &    896.67$\pm$118.77    &    1316.67$\pm$4.71    &    973.33$\pm$70.4    &    1610.00$\pm$28.58    &    1183.33$\pm$48.36    \\
12        &    12500.00$\pm$2753.48    &    16533.33$\pm$3376.96    &    17166.67$\pm$3805.55    &    22300.00$\pm$4375.50    &    16900.00$\pm$3191.39    \\
13        &    133.33$\pm$23.57    &    33.33$\pm$23.57    &    150.00$\pm$20.41    &    250.00$\pm$20.41    &    200.00$\pm$61.24    \\
14        &    3589.00$\pm$954.25    &    1008.67$\pm$184.28    &    2112.00$\pm$189.90    &    1075.00$\pm$284.00    &    4041.33$\pm$1629.86    \\
15        &    116.67$\pm$21.25    &    250.00$\pm$10.21    &    8.33$\pm$5.89    &    250.00$\pm$46.77    &    183.33$\pm$59.80    \\
16        &    -4.00$\pm$0.82    &    -8.00$\pm$1.41    &    -6.67$\pm$0.47    &    -4.00$\pm$1.41    &    -15.33$\pm$0.94    \\
17        &    133.33$\pm$47.14    &    1766.67$\pm$84.99    &    1233.33$\pm$23.57    &    866.67$\pm$188.56    &    1733.33$\pm$62.36    \\
18        &    1096.67$\pm$142.73    &    1476.67$\pm$257.56    &    1070.00$\pm$441.27    &    1010.00$\pm$102.55    &    1220.0$\pm$236.22    \\
19        &    0.00$\pm$0.00    &    0.00$\pm$0.00    &    0.00$\pm$0.00    &    0.00$\pm$0.00    &    0.00$\pm$0.00    \\
20        &    0.00$\pm$0.00    &    0.00$\pm$0.00    &    0.00$\pm$0.00    &    0.00$\pm$0.00    &    0.00$\pm$0.00    \\
21        &    116.67$\pm$31.18    &    133.33$\pm$31.18    &    266.67$\pm$51.37    &    283.33$\pm$94.28    &    166.67$\pm$47.14    \\
22        &    22.00$\pm$5.12    &    2.67$\pm$0.85    &    12.33$\pm$5.04    &    25.00$\pm$3.89    &    22.00$\pm$3.56    \\
23        &    666.67$\pm$471.4    &    1933.33$\pm$579.27    &    1800.00$\pm$561.25    &    1366.67$\pm$306.41    &    1566.67$\pm$526.52    \\
24        &    3.67$\pm$0.62    &    3.67$\pm$0.62    &    3.00$\pm$0.41    &    1.67$\pm$0.85    &    2.00$\pm$0.00    \\
25        &    713.33$\pm$54.37    &    913.33$\pm$143.84    &    486.67$\pm$23.57    &    953.33$\pm$250.38    &    506.67$\pm$106.56    \\
26        &    633.33$\pm$84.98    &    1833.33$\pm$730.68    &    4066.67$\pm$23.57    &    3066.67$\pm$731.82    &    833.33$\pm$23.57    \\
27        &    396.0$\pm$35.93    &    454.67$\pm$57.74    &    293.33$\pm$41.48    &    264.00$\pm$35.93    &    264.00$\pm$62.23    \\
28        &    2.00$\pm$0.00    &    9.33$\pm$1.17    &    16.33$\pm$2.05    &    6.00$\pm$1.87    &    41.00$\pm$7.79    \\
29        &    866.67$\pm$188.56    &    600.00$\pm$40.82    &    833.33$\pm$201.38    &    700.00$\pm$0.00    &    800.00$\pm$108.01    \\
30        &    816.67$\pm$103.79    &    2573.33$\pm$532.80    &    4016.67$\pm$671.62    &    2293.33$\pm$341.62    &    3590.00$\pm$408.84    \\
31        &    6.67$\pm$4.40    &    16.33$\pm$1.03    &    26.67$\pm$12.04    &    25.33$\pm$7.19    &    34.33$\pm$11.30    \\
32        &    4492.67$\pm$507.43    &    5942.33$\pm$338.77    &    6052.67$\pm$1475.40    &    4207.33$\pm$348.97    &    6467.67$\pm$773.16    \\
33        &    503.33$\pm$56.32    &    493.33$\pm$96.47    &    470.00$\pm$89.81    &    666.67$\pm$164.13    &    566.67$\pm$125.19    \\
34        &    -8495.67$\pm$6.33    &    -8498.33$\pm$6.80    &    -8500.67$\pm$4.50    &    -11818.33$\pm$6.34    &    -8496.67$\pm$6.98    \\
35        &    4433.33$\pm$524.93    &    11733.33$\pm$2124.59    &    9666.67$\pm$2484.73    &    16066.67$\pm$3134.31    &    6833.33$\pm$1748.49    \\
36        &    392.00$\pm$43.99    &    329.00$\pm$43.15    &    581.00$\pm$58.36    &    336.00$\pm$59.4    &    378.00$\pm$22.68    \\
37        &    0.00$\pm$0.00    &    100.00$\pm$37.41   &    246.67$\pm$66.50    &    326.67$\pm$107.81    &    320.00$\pm$28.28    \\
\bottomrule
\end{tabular}
}
\vspace{1ex}
\caption{ID raw scores of our model and baselines}
\end{table}

\begin{table}[!htbp]
\centering
\resizebox{0.88\linewidth}{!}{
\begin{tabular}{c|ccccc}
\toprule
\textbf{G} & \textbf{DT}  & \textbf{DTL}  & \textbf{DTV}  & \textbf{DTGI-a} & \textbf{DTGI}  \\ \midrule
1        &    -766.67$\pm$164.99    &    -368.33$\pm$178.53    &    -289.00$\pm$110.57    &    100.00$\pm$0.00    &    -533.33$\pm$205.48    \\
2        &    0.00$\pm$0.00    &    0.33$\pm$0.23    &    1.33$\pm$0.47    &    1.67$\pm$1.18    &    1.33$\pm$0.47    \\
3        &    -3.33$\pm$1.70    &    -6.00$\pm$0.00    &    -7.67$\pm$1.03    &    -9.33$\pm$1.31    &    -2.00$\pm$0.82    \\
4        &    0.00$\pm$0.00    &    25.00$\pm$17.68    &    0.00$\pm$0.00    &    0.00$\pm$0.00    &    50.00$\pm$17.68    \\
5        &    320.00$\pm$40.82    &    100.00$\pm$16.32    &    33.33$\pm$9.43    &    106.67$\pm$9.43    &    166.67$\pm$9.43    \\
6        &    150.0$\pm$18.71    &    143.33$\pm$20.54    &    110.00$\pm$18.71    &    143.33$\pm$18.41    &    110.0$\pm$18.71    \\
7        &    83.33$\pm$11.18    &    66.67$\pm$11.18    &    166.67$\pm$22.49    &    66.67$\pm$47.14    &    33.33$\pm$13.57    \\
8        &    66.67$\pm$20.07    &    240.00$\pm$20.41    &    26.67$\pm$15.46    &    45.00$\pm$10.61    &    146.67$\pm$27.04    \\
9        &    0.00$\pm$0.00    &    33.33$\pm$14.34    &    40.00$\pm$0.00    &    33.33$\pm$4.71    &    36.67$\pm$6.24    \\
10        &    1783.33$\pm$31.18    &    966.67$\pm$286.02    &    866.67$\pm$405.35    &    1616.67$\pm$116.07    &    1916.67$\pm$589.61    \\ 
\bottomrule
\end{tabular}
}
\vspace{1ex}
\caption{OOD raw scores of our model and baselines}
\end{table}

\clearpage
2. \textbf{Raw Scores in the Experiment 5.4}
\begin{table}[H]
\centering
\resizebox{0.5\linewidth}{!}{
\begin{tabular}{c|ccc}
\toprule
\textbf{G} & \textbf{10 Games(25\%)}  & \textbf{20 Games(50\%)}   & \textbf{37 Games(100\%)}    \\ 
\midrule
1        &    33.33$\pm$23.57   &    -429.33$\pm$225.01    &    -533.33$\pm$205.48     \\
2        &    0.00$\pm$0.00    &    0.00$\pm$0.00    &    1.33$\pm$0.47    \\
3        &     -5.00$\pm$0.00    &    -5.00$\pm$1.08    & -2.00$\pm$0.82    \\
4        &    0.00$\pm$0.00    &    483.33$\pm$341.77    &    50.00$\pm$17.68    \\
5        &    133.33$\pm$9.43    &    140.00$\pm$14.14    &   166.67$\pm$9.43   \\
6        &    40.00$\pm$0.00    &    90.00$\pm$21.60    &    110.00$\pm$18.71    \\
7        &    216.67$\pm$51.37    &    116.67$\pm$11.79    &    33.33$\pm$13.57    \\
8        &    6.67$\pm$2.36     &    0.00$\pm$0.00    &  146.67$\pm$27.04     \\
9        &    3.33$\pm$2.36    &    23.33$\pm$2.35    &    36.67$\pm$6.24    \\
10        &     400.0$\pm$61.24    &    2333.33$\pm$243.53    &    1916.67$\pm$589.61    \\ 
\bottomrule
\end{tabular}
}
\vspace{1ex}
\caption{OOD raw scores of DTGI are acquired using offline datasets comprising 10, 20, and 37 games.}
\label{table_ood_raw}
\end{table}

3. \textbf{Raw Scores in the Experiment 5.4}
\begin{table}[H]
\centering
\resizebox{0.8\linewidth}{!}{
\begin{tabular}{c|cccccc}
\toprule
\textbf{G} & \textbf{DT(10k)}  & \textbf{DT(100k)}  & \textbf{DT(200k)} & \textbf{DTGI(10k)}  & \textbf{DTGI(100k)}  & \textbf{DTGI(200k)}   \\ 
\midrule
1        &    -25900.00$\pm$5799.57    &    -15500.00$\pm$2108.32    &    -16700.00$\pm$708.28    &    -17066.67$\pm$3860.12    &    -24300.00$\pm$3149.87    &    -10333.33$\pm$4326.92    \\
2        &    2406.67$\pm$320.16    &    2370.00$\pm$104.24    &    1210.00$\pm$81.96    &    3206.67$\pm$151.35    &    3653.33$\pm$22.48    &    1820.00$\pm$364.85    \\
3        &    0.00$\pm$0.00    &    0.00$\pm$0.00    &    0.00$\pm$0.00    &    0.00$\pm$0.00    &    0.00$\pm$0.00    &    0.00$\pm$0.00    \\
4        &    12.33$\pm$2.46    &    19.00$\pm$1.87    &    19.00$\pm$0.71    &    16.00$\pm$0.71    &    18.33$\pm$0.47    &    18.00$\pm$0.00    \\
5        &    0.00$\pm$0.00    &    340.00$\pm$233.38    &    0.00$\pm$0.00    &    486.67$\pm$344.13    &    0.00$\pm$0.00    &    0.00$\pm$0.00    \\
6        &    -64.00$\pm$4.97    &    -37.67$\pm$3.65    &    -22.67$\pm$1.25    &    -58.00$\pm$1.63    &    -42.0$\pm$9.42    &    -30.67$\pm$2.36    \\
7        &    173.33$\pm$30.57    &    300.00$\pm$32.08    &    423.33$\pm$89.68    &    341.67$\pm$126.12    &    443.33$\pm$57.09    &    563.33$\pm$86.63    \\
8        &    666.67$\pm$235.70    &    6666.67$\pm$1840.90    &    7333.33$\pm$1027.40    &    5666.67$\pm$1178.51    &    3333.33$\pm$623.61    &    4000.0$\pm$707.11    \\
9        &    533.33$\pm$47.14    &    800.00$\pm$141.42    &    733.33$\pm$47.14    &    666.67$\pm$47.14    &    733.33$\pm$47.14    &    1200.00$\pm$216.02    \\
10        &    160.00$\pm$35.59    &    320.00$\pm$14.14    &    360.00$\pm$61.64    &    266.67$\pm$41.90    &    326.67$\pm$54.37    &    460.00$\pm$107.08    \\
11        &    896.67$\pm$118.77    &    803.33$\pm$12.47    &    1070.00$\pm$100.33    &    1183.33$\pm$48.36    &    1133.33$\pm$50.72    &    1243.33$\pm$165.19    \\
12        &    12500.00$\pm$2753.48    &    9433.33$\pm$1230.40    &    13466.67$\pm$2481.38    &    16900.00$\pm$3191.39    &    16900.00$\pm$1309.58    &    9766.67$\pm$2703.19    \\
13        &    133.33$\pm$23.57    &    333.33$\pm$47.14    &    200.00$\pm$20.41    &    200.00$\pm$61.24    &    350.00$\pm$35.36    &    233.33$\pm$42.49    \\    
14        &    3589.00$\pm$954.25    &    1612.33$\pm$322.67    &    3144.67$\pm$1033.14    &    4041.33$\pm$1629.86    &    2945.33$\pm$587.40    &    2129.33$\pm$611.92    \\
15        &    116.67$\pm$21.25    &    158.33$\pm$35.84    &    325.00$\pm$30.62    &    183.33$\pm$59.80    &    625.00$\pm$20.41    &    416.67$\pm$42.49    \\    
16        &    -4.00$\pm$0.82    &    -9.33$\pm$2.49    &    -2.67$\pm$1.25    &    -15.33$\pm$0.94    &    -10.67$\pm$1.25    &    -14.0$\pm$0.82    \\
17        &    133.33$\pm$47.14    &    1800.00$\pm$40.82    &    2100.00$\pm$362.86    &    1733.33$\pm$62.36    &    3633.33$\pm$836.99    &    5533.33$\pm$651.07    \\
18        &    1096.67$\pm$142.73    &    2063.33$\pm$188.13    &    2083.33$\pm$86.63    &    1220.00$\pm$236.22    &    2670.00$\pm$223.76    &    2073.33$\pm$275.75    \\
19        &    0.00$\pm$0.00    &    0.00$\pm$0.00    &    0.00$\pm$0.00    &    0.00$\pm$0.00    &    0.00$\pm$0.00    &    0.00$\pm$0.00    \\
19        &    0.00$\pm$0.00    &    0.00$\pm$0.00    &    0.00$\pm$0.00    &    0.00$\pm$0.00    &    0.00$\pm$0.00    &    0.00$\pm$0.00    \\
21        &    116.67$\pm$31.18    &    200.00$\pm$54.00    &    200.0$\pm$54.01    &    166.67$\pm$47.14    &    433.33$\pm$77.28    &    250.00$\pm$35.36    \\     
22        &    22.00$\pm$5.12    &    20.33$\pm$4.18    &    8.67$\pm$2.66    &    22.00$\pm$3.56    &    26.33$\pm$3.06    &    24.0$\pm$4.32    \\
23        &    666.67$\pm$471.4    &    866.67$\pm$117.85    &    533.33$\pm$117.85    &    1566.67$\pm$526.52    &    1233.33$\pm$224.85    &    2833.33$\pm$946.34    \\
24        &    3.67$\pm$0.62    &    5.67$\pm$0.23    &    6.00$\pm$0.41    &    2.00$\pm$0.00    &    5.67$\pm$1.03    &    3.00$\pm$1.08    \\
25        &    713.33$\pm$54.37    &    426.67$\pm$77.17    &    773.33$\pm$54.37    &    506.67$\pm$106.56    &    713.33$\pm$70.40    &    753.33$\pm$141.97    \\
26        &    633.33$\pm$84.98    &    2833.33$\pm$789.87    &    2933.33$\pm$824.96    &    833.33$\pm$23.57    &    1966.67$\pm$758.65    &    2866.67$\pm$805.54    \\
27        &    396.00$\pm$35.93    &    293.33$\pm$51.85    &    542.67$\pm$27.44    &    264.00$\pm$62.23    &    278.67$\pm$27.44    &    572.00$\pm$35.93    \\    
28        &    2.00$\pm$0.00    &    38.67$\pm$10.74    &    11.67$\pm$4.73    &    41.00$\pm$7.79    &    50.67$\pm$14.27    &    35.00$\pm$13.83    \\
29        &    866.67$\pm$188.56    &    733.33$\pm$23.57   &    1233.33$\pm$143.37    &    800.0$\pm$108.01    &    1100.0$\pm$177.95    &    2533.33$\pm$601.85    \\
30        &    816.67$\pm$103.79    &    3530.00$\pm$541.03    &    3590.00$\pm$723.06    &    3590.00$\pm$408.84    &    3213.33$\pm$895.29    &    5496.67$\pm$896.96    \\
31        &    6.67$\pm$4.40    &    56.0$\pm$6.75    &    64.00$\pm$5.31    &    34.33$\pm$11.3    &    71.33$\pm$5.95    &    70.33$\pm$3.68    \\
32        &    4492.67$\pm$507.43    &    4581.33$\pm$125.08    &    6988.33$\pm$1482.60    &    6467.67$\pm$773.16    &    3226.67$\pm$450.93    &    8352.00$\pm$2537.06    \\
33        &    503.33$\pm$56.32    &    796.67$\pm$139.06    &    656.67$\pm$27.18    &    566.67$\pm$125.19    &    503.33$\pm$20.14    &    993.33$\pm$290.10    \\
34        &    -8495.67$\pm$6.33    &    -8488.67$\pm$5.57    &    -8491.00$\pm$5.12    &    -8496.67$\pm$6.98    &    -8484.00$\pm$4.71    &    -11253.00$\pm$1950.20    \\
35        &    4433.33$\pm$524.93    &    6866.67$\pm$1569.15    &    6533.33$\pm$1821.32    &    6833.33$\pm$1748.49    &    17800.00$\pm$1568.97    &    10566.67$\pm$410.96    \\
36        &    392.00$\pm$43.99    &    427.00$\pm$4.95    &    434.00$\pm$49.50    &    378.00$\pm$22.68    &    511.00$\pm$38.66    &    539.00$\pm$40.52    \\
37        &    0.00$\pm$0.00    &    313.33$\pm$188.86    &    106.67$\pm$24.94    &    320.00$\pm$28.28    &    160.00$\pm$37.42    &    380.00$\pm$108.01    \\
\bottomrule
\end{tabular}
}
\vspace{1ex}
\caption{ID raw scores of our model and DT, utilizing varying sizes of offline datasets.}
\label{table_id_raw_1}
\end{table}

\begin{table}[H]
\centering
\resizebox{0.8\linewidth}{!}{
\begin{tabular}{c|cccccc}
\toprule
\textbf{G} & \textbf{DT(10k)}  & \textbf{DT(100k)}  & \textbf{DT(200k)} & \textbf{DTGI(10k)}  & \textbf{DTGI(100k)}  & \textbf{DTGI(200k)}   \\ 
\midrule
1        &    -766.67$\pm$164.99    &    -1000.00$\pm$-0.00    &    4083.67$\pm$3559.4    &    -533.33$\pm$205.48    &    -405.67$\pm$214.75    &    -633.33$\pm$224.85    \\
2        &    0.00$\pm$0.00    &    2.33$\pm$2.33    &    0.00$\pm$0.00    &    1.33$\pm$0.47    &    2.00$\pm$0.00    &    2.33$\pm$0.24    \\
3        &    -3.33$\pm$1.70    &    -7.00$\pm$1.08    &    -7.00$\pm$1.08    &    -2.00$\pm$0.82    &    -2.00$\pm$0.00    &    -7.67$\pm$0.94    \\
4        &    0.00$\pm$0.00    &    0.00$\pm$0.00    &    0.00$\pm$0.00    &    50.00$\pm$17.68    &    75.00$\pm$0.00    &    75.00$\pm$0.00    \\
5        &    320.00$\pm$40.82    &    166.67$\pm$41.10    &    153.33$\pm$18.86    &    166.67$\pm$9.43    &    86.67$\pm$20.55    &    140.00$\pm$37.42    \\
6        &    150.00$\pm$18.71    &    123.33$\pm$15.46    &    76.67$\pm$4.71    &    110.00$\pm$18.71    &    43.33$\pm$6.24    &    76.67$\pm$6.24    \\
7        &    83.33$\pm$31.18    &    116.67$\pm$23.57    &    100.00$\pm$20.41    &    33.33$\pm$23.57    &    166.67$\pm$11.79    &    166.67$\pm$31.18    \\
8        &    66.67$\pm$40.07    &    68.33$\pm$16.62    &    226.67$\pm$31.64    &    146.67$\pm$37.04    &    111.67$\pm$8.25    &    171.67$\pm$75.45    \\
9        &    0.00$\pm$0.00    &    60.00$\pm$10.80    &    36.67$\pm$14.34    &    36.67$\pm$6.24    &    23.33$\pm$6.24    &    70.00$\pm$22.73    \\
10        &    1783.33$\pm$31.18    &    2733.33$\pm$295.33    &    750.00$\pm$196.85    &    1916.67$\pm$589.61    &    2550.00$\pm$636.40    &    1916.67$\pm$307.09    \\ 
\bottomrule
\end{tabular}
}
\vspace{1ex}
\caption{OOD raw scores of our model and DT, utilizing varying sizes of offline datasets.}
\label{table_ood_raw_2}
\end{table}

\subsection{Experimental Details}
\label{append_3}
\textbf{Training} 
To ensure fairness and consistency, we maintain uniformity across all baseline models and our model, encompassing both hyper-parameters and the training process.
The specific hyper-parameters used for our model and the baseline models are detailed in Table \ref{table_hp_model}. For training purposes, two NVIDIA 4090 graphics cards are utilized in Experiments \ref{exp_1}, \ref{exp_3}, and \ref{exp_4}, while eight NVIDIA 4090 graphics cards are employed in Experiment \ref{exp_2}.

\begin{table}[H]
\centering
\resizebox{0.5\linewidth}{!}{
\begin{tabular}{c|c}
\toprule
\textbf{hyper-parameters} & \textbf{value}   \\ 
\midrule
Context length (DT)       &    20   \\
Number of layers (DT)      &    6   \\
Number of heads (DT)         &    8      \\
embedding dim (DT)       &     128      \\
embedding dim (CLIP)       &     512      \\
input dim (hypernetworks)        &    512   \\
Number of layers ($\mathrm{Encoder_f}$ / $\mathrm{Encoder_g}$) &  1  \\
Number of heads ($\mathrm{Encoder_f}$ / $\mathrm{Encoder_g}$)  &    2 \\
Nonlinearity & ReLU, encoder; GeLU, otherwise \\
Dropout & 0.1 \\
Max\_epochs &  30 \\
Batch\_size & 1000, experiment \ref{exp_2}; 200, otherwise    \\
Learning\_rate & 6e-4 \\
Betas & (0.9, 0.95) \\
Grad\_norm\_clip & 1.0 \\
Weight\_decay & 0.1 \\
Warmup\_tokens & 512*20 \\
optimizer & AdamW \\
\bottomrule
\end{tabular}
}
\vspace{1ex}
\caption{Hyper-parameters.}
\label{table_hp_model}
\end{table}

\textbf{Evaluation} 
We evaluate each model three times per game from 3 different seeds, presenting the mean and the standard deviation. 
The maximum steps per evaluation is 5120.
The targeted return is set at 500 for all games.
Each model is evaluated thrice per game, using three distinct seeds, and we report the mean and std. 
The maximum number of steps per evaluation is set to 5120. 
The target return is uniformly set at 500 for all games.

\textbf{Game Instruction} We construct a multimodal game instruction set. Some properties of this instruction set are listed in Table \ref{table_hp_model_2}.
\begin{table}[H]
\centering
\resizebox{0.5\linewidth}{!}{
\begin{tabular}{c|c}
\toprule
\textbf{properties} & \textbf{value}   \\ 
\midrule
Number of games & 47 \\
Number of instructions & 47*50 \\
Contents in a instruction & description, trajectory, guidance \\
Length of trajectory in a instruction & 20 \\
\bottomrule
\end{tabular}
}
\vspace{1ex}
\caption{properties of the instruction set}
\label{table_hp_model_2}
\end{table}

\clearpage
\subsection{instruction Important Scores for All Games}
\label{append_4}

\begin{figure}[H]
\centering
\includegraphics[width=0.80\linewidth]{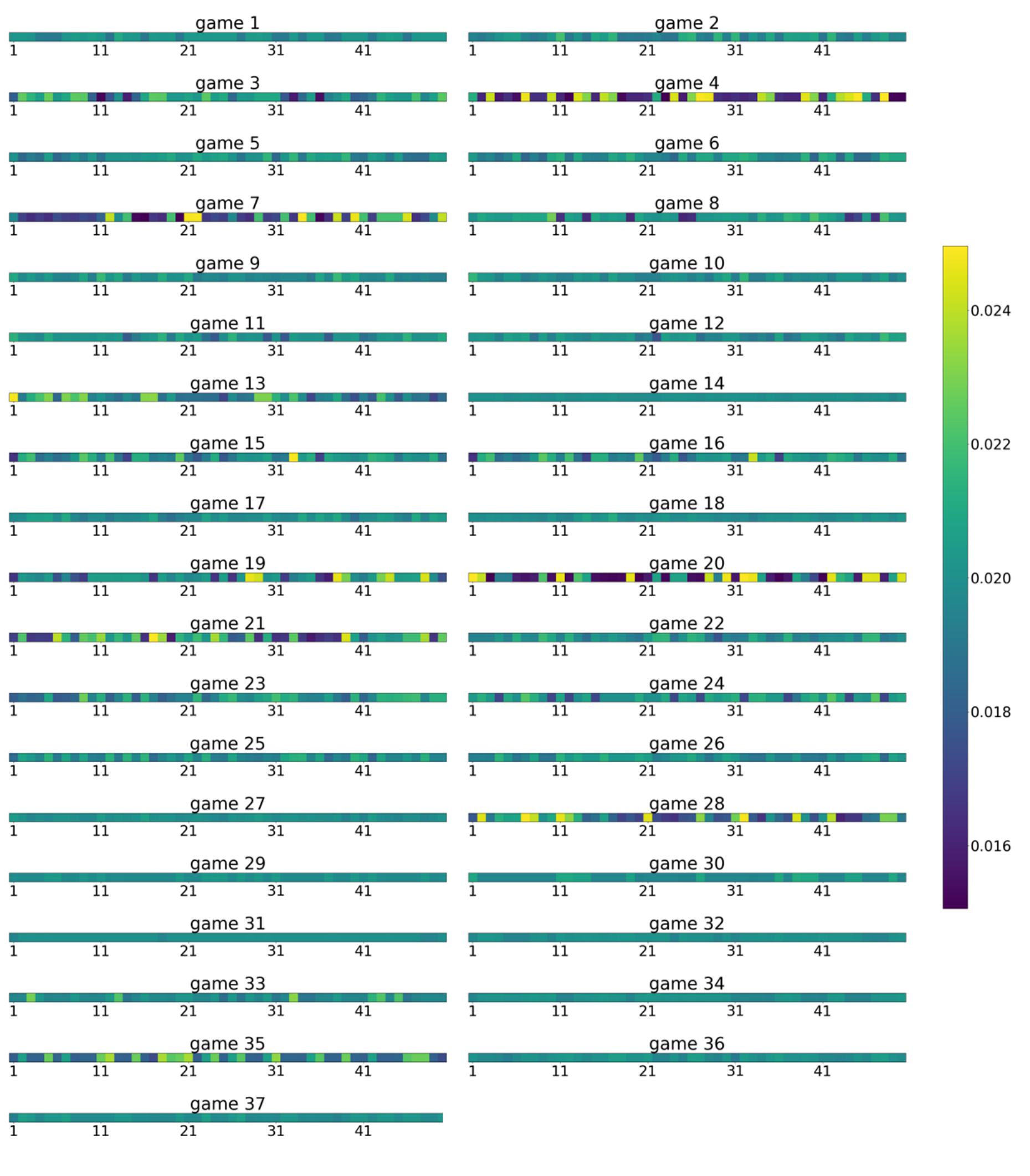}
\caption{Instruction Importance scores for 37 training games.} 
\label{fig_s1}
\end{figure}

\begin{figure}[H]
\centering
\includegraphics[width=0.80\linewidth]{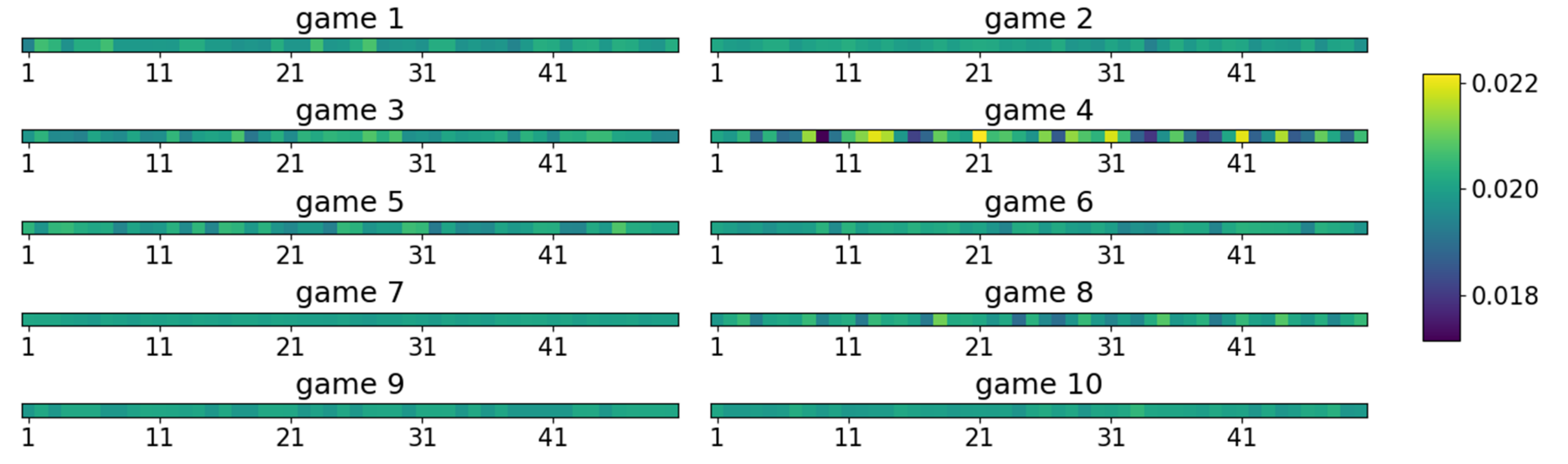}
\caption{Instruction Importance scores for 10 unseen games.} 
\label{fig_s2}
\end{figure}

\end{document}